\documentclass[authoryear,preprint,review,12pt]{elsarticle}

\usepackage{amssymb}
\usepackage{lineno}

\usepackage{adjustbox}
\usepackage{tikz}
\usepackage{tabularx}

\usepackage{caption}
\usepackage{subcaption}
\usepackage{amssymb}
\usepackage{tikz}
\usepackage{lineno}
\usepackage{fancyhdr,lipsum}
\usepackage{setspace}
\usepackage{xspace}
\usepackage{amssymb}
\usepackage{booktabs}
\usepackage{diagbox}
\usepackage{graphicx}
\usepackage{float} 
\usepackage{caption}
\usepackage{parskip}
\usepackage{amsmath}
\usepackage{multirow}
\usepackage{makecell}
\usepackage{algorithm}
\usepackage{algpseudocode}
\usepackage{amsthm}
\usepackage{appendix}
\usepackage{makecell}
\usepackage{bm}

\floatstyle{plaintop}
\restylefloat{table}
\usepackage[colorlinks,
            linkcolor=blue,
            anchorcolor=blue,
            citecolor=blue
            ]{hyperref}
            
\journal{Remote Sensing of Environment}

\makeatletter
\newcommand*{\etc}{%
    \@ifnextchar{.}%
        {etc}%
        {etc.\@\xspace}%
}

\begin{document}

\begin{frontmatter}



\title{Knowledge-guided machine learning for county-level corn yield prediction under drought}


\author[inst1]{Xiaoyu Wang}
\ead{xwang2696@wisc.edu}
\author[inst1]{Yijia Xu}
\ead{xu556@wisc.com}
\author[inst2]{Jingyi Huang}
\ead{jhuang426@wisc.edu}
\author[inst3]{Zhengwei Yang}
\ead{zhengwei.yang@usda.gov}
\author[inst4]{Yanbo Huang}
\ead{yanbo.huang@usda.gov}
\author[inst5]{Rajat Bindlish}
\ead{rajat.bindlish@nasa.gov}
\author[inst1]{Zhou Zhang\corref{cor1}}
\ead{zzhang347@wisc.edu}
\cortext[cor1]{Corresponding author}

\affiliation[inst1]{organization={Biological Systems Engineering},%Department and Organization
            addressline={University of Wisconsin-Madison}, 
            city={Madison},
            postcode={53706}, 
            state={WI},
            country={USA}}

\affiliation[inst2]{organization={Department of  of Soil and Environmental Sciences},%Department and Organization
            addressline={University of Wisconsin-Madison}, 
            city={Madison},
            postcode={53706}, 
            state={WI},
            country={USA}}

\affiliation[inst3]{organization={U.S. Department of Agriculture, National Agricultural Statistics Service}, 
            city={Washington}, 
            postcode={20250}, 
            state={DC}, 
            country={USA}}

\affiliation[inst4]{organization={U.S. Department of Agriculture, Agricultural Research Service, Genetics and Sustainable Agricultural Research Unit}, 
            city={Mississippi State}, 
            postcode={39762}, 
            state={MS}, 
            country={USA}}

\affiliation[inst5]{organization={NASA Goddard Space Flight Center}, 
            city={Greenbelt}, 
            postcode={20771}, 
            state={MD}, 
            country={USA}}

\begin{abstract}

Remote sensing (RS) provides large-scale, non-contact observations that are valuable for crop yield prediction. Process-based models are based on crop growth mechanisms but often struggle with large RS datasets and require heavy calibration. Machine learning (ML) models can use RS data effectively but are often criticized as “black boxes” with low interpretability. To overcome these issues, we proposed the Knowledge-Guided Machine Learning with Soil Moisture (KGML-SM) framework, which combined the strengths of both approaches. Soil moisture was used as an intermediate variable, linking weather and crop growth and serving as a key factor in drought impacts on yield. This design improved interpretability by tracing yield prediction errors to soil moisture estimates. We also added a drought-aware loss function that penalized overestimation in dry regions, making the model more robust under drought stress. This study focused on the U.S. Corn Belt, covering 12 states and more than 800 counties from 2019 to 2023. We constructed two datasets: a field-level dataset generated from the Agricultural Production Systems sIMulator (APSIM) and a county-level dataset built with satellite-based MODIS RS data and gridded climate data. Model performance was evaluated against USDA-NASS county-level yield records. KGML-SM achieved lower errors than traditional ML baselines, with an RMSE of 1.071 t/ha and an $R^2$ of 0.807 in 2023. Attention-based analysis further revealed the role of drought and soil moisture in yield prediction. Overall, KGML-SM improves both accuracy and interpretability, offering insights for future model development and climate-resilient agriculture.

\end{abstract}


\end{frontmatter}

% \linenumbers

\section{Introduction}

Corn, as a primary crop, plays a vital role in U.S. agriculture, supporting food security as well as the animal feed and biofuel industries \citep{corn_importance_1, corn_importance_3}. Accurate yield prediction is crucial for effective resource management and economic stability \citep{corn_importance_2}. Nevertheless, achieving accurate prediction across large areas and diverse environmental conditions remains challenging \citep{scym}. Climate extremes such as drought pose a major obstacle to accuracy, particularly at the county level where heterogeneous management practices and soil conditions introduce additional variability \citep{review_4_2}.

Traditional process-based models are grounded in a detailed understanding of the physical, chemical, and biological processes of crop growth \citep{process_1}. These models use equations derived from scientific principles to simulate crop, weather, and soil interactions, which makes them highly interpretable and reliable when underlying processes are well characterized \citep{process_2}. Prominent frameworks include the Agricultural Production Systems sIMulator (APSIM) \citep{apsim}, the Decision Support System for Agrotechnology Transfer (DSSAT) \citep{dssat}, the Agricultural Policy/Environmental eXtender (APEX) \citep{apex}, and the Ecosys Modelling Project (ecosys) \citep{ecosys}, all of which have been successfully applied to crop yield prediction \citep{process_3, process_4, process_5, process_6}. Yet plant growth is highly complex, and the reliance on a limited set of fixed inputs prevents these models from fully capturing this complexity \citep{scym, process_7}. In addition, they require extensive manual parameter tuning, which constrains their use across large regions \citep{process_0, process_4}.

In recent years, remote sensing (RS) has provided several benefits for corn yield prediction \citep{scym}. It enables the collection of large-scale real-time data across vast agricultural areas. These data, typically combined with machine learning (ML), facilitate accurate and efficient large-scale yield prediction. Deep learning (DL), a subset of ML, is based on neural networks and is particularly effective for high-dimensional RS data, as it can automatically extract features from large datasets and capture complex environment–yield relationships without manual engineering \citep{dlbook, dl_1}. Several studies have demonstrated the potential of DL for yield prediction, such as scalable representation learning \citep{rsml_research_1}, transfer learning across regions \citep{rsml_research_2}, and county-level prediction using Bayesian neural networks \citep{rsml_research_3}. More recently, adaptive multi-modal fusion frameworks have been proposed to integrate heterogeneous data sources and provide improved accuracy \citep{review_2_1}. However, DL methods face important limitations: they require large amounts of training data, often function as black boxes with limited interpretability, and struggle to represent the biological processes linking soil, weather, and crop growth. In contrast, process-based models explicitly capture these mechanisms but are less suited for large-scale RS integration. This complementarity highlights the need for approaches that combine the strengths of both, motivating recent research on combining these two kinds of models. A recent study \citep{review_4_4} also emphasized that relying solely on either process-based or ML models is insufficient for robust yield prediction. This growing recognition of complementarity has motivated the emergence of knowledge-guided machine learning (KGML), which seeks to integrate process-based understanding into ML frameworks to enhance both accuracy and interpretability.

KGML \citep{kgml} aims to integrate scientific knowledge into ML frameworks to achieve better performance, scientific consistency, and explainability of results. This paradigm involves three main approaches: (i) knowledge-guided learning, which incorporates scientific laws into algorithms through modified loss functions \citep{kgml_10, kgml_11}; (ii) knowledge-guided architectures, which embed knowledge directly into model structures \citep{kgml_12, kgml_13}; and (iii) knowledge-guided pretraining, which uses simulated data or self-supervised tasks to initialize models \citep{kgml_8, kgml_9}. In crop yield prediction, most KGML applications rely on DL rather than traditional ML approaches \citep{kgml_crop_2, kgml_crop_3, kgml_crop_4}. Since process-based models already provide insights into crop growth and generate high-quality simulated data, Knowledge-Guided Pretraining has become the most popular strategy. These studies typically use ML to learn yield-relevant patterns while incorporating simulated intermediates as mechanistic links from weather and soil to crop growth \citep{kgml_crop_3, kgml_crop_4, kgml_crop_5}. Specifically, \citet{kgml_crop_3} applied physics-guided reweighting with ecosystem-process intermediates—ecosystem autotrophic respiration (Ra), ecosystem heterotrophic respiration (Rh), and net ecosystem exchange (NEE)—to improve time-aware robustness and interpretability. \citet{kgml_crop_4} coupled process-model surrogates with Ensemble Kalman Filter (EnKF) data assimilation to fuse leaf area index (LAI), gross primary production (GPP), and evapotranspiration (ET) for yield prediction. \citet{kgml_crop_5} linked image-derived plant traits to a knowledge-guided S-shaped growth curve to forecast fruit growth, yield, and maturity. Recently, \citet{review_2_2} developed a physics-informed recurrent neural network for yield loss forecasting, where crop water use (ETa) and drought sensitivity (Ky) were estimated under physical constraints.

% why study soil moisture:  KGML limitation 
Despite these advances, no study has explicitly embedded soil moisture into the ML model structure. Soil moisture is central to crop production because it mediates the effects of weather on plant growth and yield formation \citep{drought_corn_1}. Unlike vegetation indices (VIs) which capture plant status at a single moment, soil moisture reflects the cumulative effects of precipitation (PPT) and temperature over time, making it a more process-oriented indicator of water availability. Embedding soil moisture explicitly in a KGML model not only provides a mechanistic link between weather and yield but also creates diagnostic interpretability, allowing prediction errors to be traced to potential misrepresentation of soil water dynamics.

To effectively exploit soil moisture for yield prediction, reliable large-scale observations are needed. Several soil moisture products have been developed over the past decades and provide valuable long-term global records. The ASCAT product offers C-band scatterometer-based soil moisture observations \citep{ascat}. In addition to satellite scatterometer data, reanalysis products such as ERA5-Land \citep{ERA5} and MERRA-2 \citep{MERRA} provide surface and root-zone soil moisture at global scales. Furthermore, the ESA product delivers more than 30 years of consistent global daily soil moisture records by merging multiple satellite sensors \citep{esa}. Building on these advances, the Soil Moisture Active Passive (SMAP) mission \citep{smap1, smap2} delivers high-accuracy global soil moisture observations with a revisit cycle of two to three days. Compared with earlier C- or X-band missions, SMAP’s L-band measurements penetrate deeper into the soil and are less affected by vegetation and atmospheric disturbances, making it one of the most reliable datasets for agricultural applications.

% research about soil moisture and yield prediction
Previous studies consistently demonstrate a strong link between soil moisture, corn yield, and drought impacts. For instance, \citet{review_4_1} showed that assimilating remotely sensed soil moisture and VIs into a crop simulation model substantially improved corn yield prediction. In addition, \citet{review_4_2} reported that the combined influence of soil moisture and atmospheric evaporative demand explained most of the observed interannual variability in U.S. corn yields, underscoring the need to jointly consider soil and atmospheric drivers. At the field scale, \citet{review_4_3} found that soil water content in deeper layers during reproductive stages showed the strongest correlation with corn yield, and that integrating high-resolution VIs improved precision irrigation decisions. Other studies have further explored related directions, such as improving retrieval accuracy by combining ASCAT and SMAP observations \citep{review_1_1}, developing cumulative drought indices (CDI) from process-based models to enhance subfield yield predictions \citep{review_1_2}, or applying ML and DL approaches for soil moisture prediction and yield modeling \citep{review_4_5, review_4_6, review_4_7}. Collectively, these works (\autoref{tab:related_work}) underscore the central role of soil moisture in corn production, but most treat it only as an auxiliary covariate or prediction target. In contrast, our KGML-SM framework embeds soil moisture explicitly as an intermediate variable, creating a mechanistic link between weather and yield and enabling diagnostic interpretability. Building on this design, KGML-SM also introduces a drought-aware loss function to penalize overestimation under water-limited conditions, thereby improving robustness. Unlike approaches that simply add soil moisture as another input feature or prediction, our framework provides a structured way to trace yield errors back to soil–weather interactions. Moreover, the proposed modules are general: embedding soil moisture as an intermediate variable can be integrated into other crop yield models to enhance interpretability, while the drought-aware loss can be applied more broadly in drought-prone regions.

% \clearpage

\begin{table}[htbp]
  \centering
  \caption{Summary of KGML applications in crops and soil moisture–related studies in context of this work.}
  \scalebox{0.75}{
    \begin{tabular}{p{2cm} p{9.6cm} p{5.2cm}}
    \toprule
    \textbf{Reference} & \textbf{Content} & \textbf{Method} \\
    \midrule
    \citep{kgml_crop_3} &
    Using physics-guided neural networks for time-aware fairness in crop yield prediction &
    LSTM with attention; physics-guided reweighting; fairness refinement \\
    
    \citep{kgml_crop_4} &
    Integrating process-based surrogates with multi-source data assimilation for agroecosystem prediction  &
    GRU surrogate; EnKF; data fusion \\
    
    \citep{kgml_crop_5} &
    Predicting greenhouse strawberry growth trajectory and yield via knowledge-guided computer vision &
    Faster R-CNN; DenseNet-based trait extraction; S-shaped growth-curve modeling \\
    
    \citep{scym} &
    Mapping large-area crop yield from satellites via pseudo-observations and calibration &
    Pseudo-observation conversion; regression calibration \\
    
    \citep{review_1_2} &
    Improving subfield corn yield prediction by including in-season water deficit with RS VIs  &
    SALUS-simulated CDI by yield-stability zones; Random Forest (RF) per tile-year and a two-window composite  \\
    
    \citep{review_2_1} &
    Fusing multi-modal remote-sensing data for optimal subfield yield prediction &
    Multi-modal encoders; gated fusion \\
    
    \citep{review_2_2} &
    Exploring physics-informed neural networks for crop yield loss forecasting &
    PINN with physics-based constraints \\
    
    \citep{review_4_1} &
    Assimilating RS soil moisture and vegetation into a crop model for corn yield prediction &
    EnKF with DSSAT-CSM (Maize); sequential SM/LAI assimilation \\
    
    \citep{review_1_1} &
    Integrating ASCAT and SMAP for global surface soil moisture retrieval using ML &
    Model comparison: RF / LSTM / SVM / CNN \\
    
    \citep{review_4_2} &
    Combining soil moisture and atmospheric evaporative demand to predict US corn yields &
    Statistical modeling with process-based analysis \\
    
    \citep{review_4_3} &
    Relating soil water content and high-resolution imagery for precision irrigation in corn &
    Linear regression with VIs coupling \\
    
    \citep{review_4_5} &
    Benchmarking DL vs. ML models for soil moisture prediction under irrigation treatments \\
    
    \citep{review_4_6} &
    Comparing environmental variables and ML algorithms for corn yield in the US Midwest &
    Comparative ML assessment \\
    
    \citep{review_4_7} &
    Predicting multi-depth soil water content during summer corn for irrigation planning &
    Residual bidirectional LSTM  \\
    
    This study &
    The first study to embed soil moisture explicitly as an intermediate variable within the ML architecture for county-level corn yield prediction; mitigating overestimation under drought area &
    Knowledge-guided pretraining; W2S encoder; drought-aware loss function \\
    \bottomrule
    \end{tabular}
  }
  \label{tab:related_work}
\end{table}

In summary, this study introduces a KGML-SM framework to address the limitations of existing process-based and ML approaches for county-level corn yield prediction. The framework explicitly embeds soil moisture as an intermediate variable through a Weather-to-Soil (W2S) encoder and an attention mechanism \citep{attention}, creating a mechanistic link between weather drivers and crop outcomes. In addition, a drought-aware loss function penalizes overestimation under water-limited conditions, thereby improving model robustness during drought years. By jointly enhancing accuracy, interpretability, and robustness, KGML-SM improves yield prediction. To our knowledge, this is the first framework to explicitly embed soil moisture into a KGML model for county-level prediction, offering both methodological innovation and practical relevance for supporting resilient agricultural management under increasing climate variability.

\section{Data acquisition}
\label{Data acquisition}

In this study, we developed two datasets for corn yield prediction. The first dataset was a field-level dataset generated using APSIM \citep{apsim} and used for pretraining. The second dataset was a county-level dataset derived from Google Earth Engine (GEE) \citep{gee} and USDA NASS \citep{usda_nass_quickstats}, and used for finetuning. The workflow was divided into three main steps: simulation, pretraining, and finetuning. In this section, we first introduce the study area (\autoref{Study area}); then, we provide details of the APSIM field-level dataset (\autoref{APSIM field-level dataset}); finally, we describe the construction of the GEE county-level dataset (\autoref{GEE county-level dataset}).

\subsection{Study area}
\label{Study area}

\begin{figure*}[htbp]
\centering  
\includegraphics[width=1\textwidth]{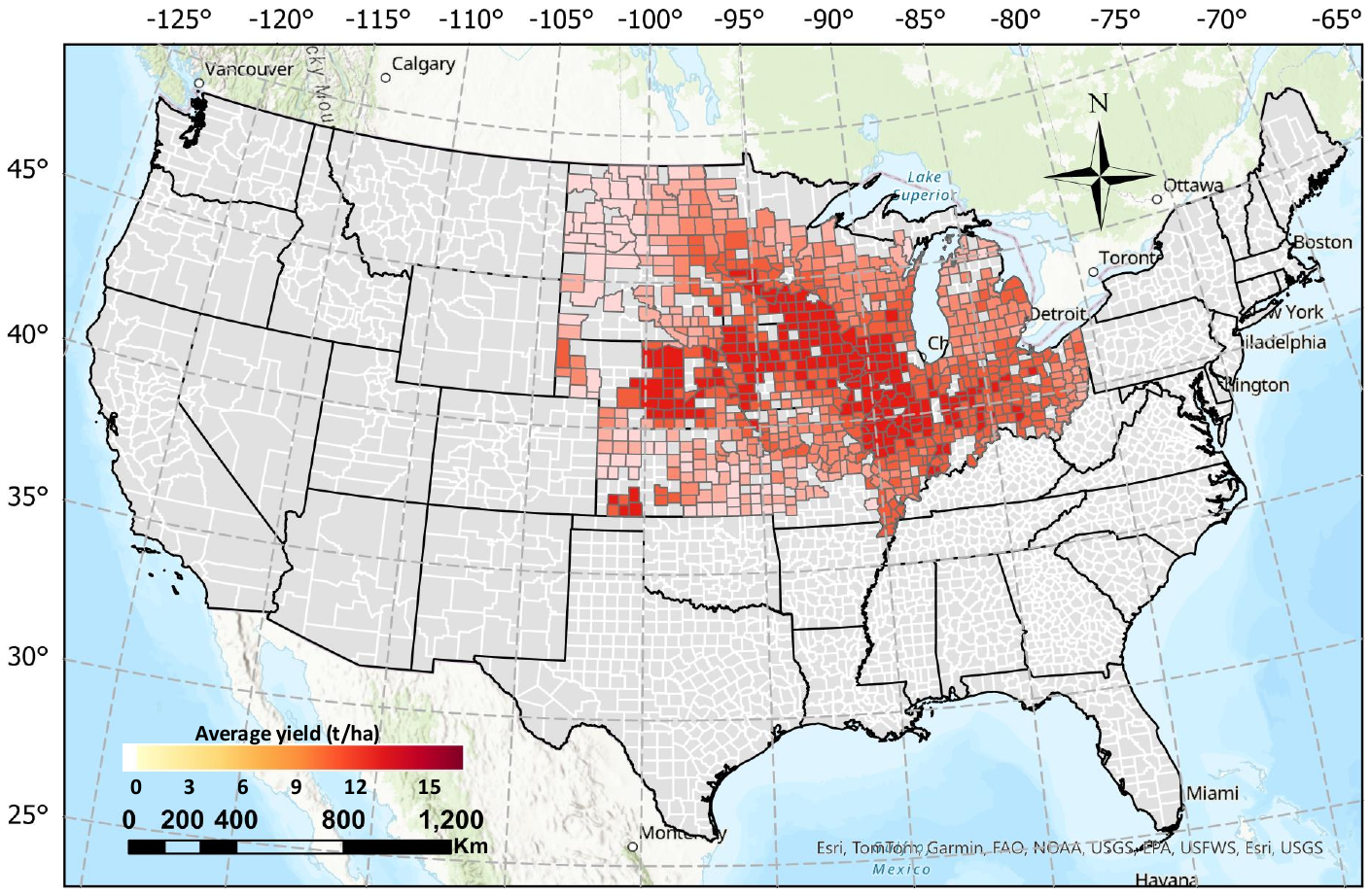}

\caption{The 5-year average county-level corn yield map in selected states.}
    \label{fulfig1}
\label{fig:studyarea}
\end{figure*}

Our research focused on corn yield prediction across the U.S. Corn Belt. Twelve states were selected as our study area, including North Dakota, South Dakota, Minnesota, Wisconsin, Iowa, Illinois, Indiana, Ohio, Missouri, Kansas, Nebraska, and Michigan. These states are crucial agricultural states in the U.S., known for their significant contributions to corn production. We generated a five-year average yield map (\autoref{fig:studyarea}) for these twelve states and considered them well suited for corn yield prediction research. For brevity, state names were referred to by their standard abbreviations (e.g., Wisconsin as WI) in the following sections.

\clearpage

\subsection{APSIM field-level dataset}
\label{APSIM field-level dataset}

In this section, we explain how APSIM was used to generate a field-level simulated dataset for model pretraining. We used daily weather data from the Iowa Environmental Mesonet (IEM) \citep{iowaiem}, a platform developed by Iowa State University that provides agricultural and environmental observations across the United States. Specifically, we extracted all available station-based records, which include variables such as temperature, PPT, wind, and solar radiation, together with the latitude and longitude of each station. This APSIM field-level dataset covers several thousand stations across 12 states from 1980 to 2023 (\autoref{tab:state_summary}).

\begin{table}[htbp]
  \centering
  \caption{The number of stations used for APSIM field-level dataset and the average number of counties used for GEE county-level dataset in each state.}
  \scalebox{0.75}{
    \begin{tabular}{lllllllllllll}
    \toprule
    \textbf{State} & IL & IN & IA & KS & MI & MN & MO & NE & ND & OH & SD & WI \\
    \midrule
    \makecell{\textbf{Station number}\\ \textbf{(APSIM field-level dataset)}} & 120 & 89 & 113 & 150 & 140 & 140 & 130 & 128 & 79 & 100 & 114 & 147 \\
    \makecell{\textbf{County number}\\ \textbf{(GEE county-level dataset)}}
    & 91 & 74 & 87 & 74 & 57 & 64 & 74 & 69 & 48 & 72 & 39 & 62 \\
    \bottomrule
    \end{tabular}%
  }
  \label{tab:state_summary}
\end{table}

\subsubsection{Input and output of APSIM simulation}
\label{Input and output of APSIM simulation}

The APSIM model simulates corn yield based on four weather data inputs: maximum and minimum temperature (Tmax and Tmin), PPT, and radiation (Radn). Tmax and Tmin influence the rate of plant development, and affect processes like photosynthesis and respiration. PPT are essential for modeling soil moisture levels, which directly affect water availability for crops. Radn is a critical factor in photosynthesis, as it provides the energy needed for plant growth. These weather data of IEM stations are available from the IEM website \citep{iowaiem}. The simulation also requires the station's latitude, longitude, and year. Some management parameters are also necessary, such as the start and end of the sowing window, plant population, fertilizer amount, and initial soil water, which were described in detail in \autoref{APSIM management parameters}. The APSIM model takes weather data as input to simulate root zone soil moisture (SM\_rootzone), surface soil moisture (SM\_surface), and corn yield. SM\_rootzone is crucial for corn growth as it directly affects water availability for uptake, influencing plant development and yield. SM\_surface plays a key role in seed germination and early growth stages.

\subsubsection{Variables summary in field-level dataset for pretraining}
\label{Variables summary in field-level dataset for pretraining}

We adopted a five-year averaging strategy consistent with prior studies \citep{rsml_research_3, att_model}, which enhances robustness by mitigating short-term variability while retaining long-term soil–climate signals. The resulting five-year historical average of simulated corn yield (Historical avg. yield) was used as a local baseline to improve model accuracy. We also included the prediction year and location (Lat\&Long) of IEM stations in the field-level dataset. Because soil properties and topography vary across space and time, year and location can help the ML model capture this variation. We then combined the weather data, simulated soil moisture and corn yield, along with other variables to construct the APSIM field-level dataset (\autoref{tab:combined_dataset}). VIs were not included in the pretraining dataset because the IEM stations, although co-located with RS imagery, are not necessarily situated in cornfields, making such signals unreliable for yield modeling. To avoid introducing noise, VIs were set to zero during pretraining, while in the finetuning stage, actual RS-based VIs were incorporated at the county level where crop type and coverage are reliably defined.

\clearpage

\begin{table}[htbp]
  \centering
  \caption{Summary of variables in the APSIM field-level dataset and the GEE county-level dataset}
  \scalebox{0.7}{
    \begin{tabular}{llcccccc}
    \toprule
    \multirow{2}{*}{\textbf{Category}} & \multirow{2}{*}{\textbf{Variables}} 
      & \multicolumn{3}{c}{\textbf{APSIM field-level dataset}} 
      & \multicolumn{3}{c}{\textbf{GEE county-level dataset}} \\
    \cmidrule(lr){3-5} \cmidrule(lr){6-8}
      &   & Unit & Type & Source & Unit & \makecell{Spatial \\ resolution} & Source \\
    \midrule
    \multirow[t]{4}{*}{Weather data} 
      & Radn & MJ/m$^2$ & Input & IEM  & W/m$^2$ & 4 km  & PRISM \\
      & Tmax & $^\circ$C & Input & IEM  & $^\circ$C & 4 km  & PRISM \\
      & Tmin & $^\circ$C & Input & IEM  & $^\circ$C & 4 km  & PRISM \\
      & PPT  & mm        & Input & IEM  & mm       & 4 km  & PRISM \\
    \multirow[t]{2}{*}{Soil moisture} 
      & SM\_surface  & -- & Simulated & APSIM & -- & 9 km   & SMAP \\
      & SM\_rootzone & -- & Simulated & APSIM & -- & 9 km   & SMAP \\
    Corn yield 
      & Yield & t/ha & Simulated & APSIM & t/ha & -- & USDA NASS \\
    \multirow[t]{4}{*}{VIs} 
      & GCVI & -- & -- & -- & -- & 500 m & MODIS \\
      & EVI      & -- & -- & -- & -- & 500 m & MODIS \\
      & NDWI     & -- & -- & -- & -- & 500 m & MODIS \\
      & NDVI     & -- & -- & -- & -- & 500 m & MODIS \\
    \multirow[t]{3}{*}{Others} 
      & Prediction year         & --        & Input     & IEM   & --      & -- & NASS \\
      & Location (Lat\&Long)    & Lat\&Long & Input     & IEM   & Lat\&Long & -- & USDA NASS \\
      & Historical avg. yield   & t/ha      & Simulated & APSIM & t/ha    & -- & USDA NASS \\
    \bottomrule
    \end{tabular}
  }
  \label{tab:combined_dataset}
\end{table}

\subsection{GEE county-level dataset}
\label{GEE county-level dataset}

In this section, the construction of a county-level dataset for finetuning is described. Details of the studied counties are provided in \autoref{tab:state_summary}. This dataset contained all variable types from the APSIM field-level dataset (\autoref{tab:combined_dataset}), including weather, soil moisture, and other features, but here they were obtained from RS and reanalysis products available in GEE. To enhance our model performance with RS data, four VIs from satellite imagery were also included: Green Chlorophyll Index (GCVI), Enhanced Vegetation Index (EVI), Normalized Difference Water Index (NDWI), and Normalized Difference Vegetation Index (NDVI). Detailed descriptions of these data sources and processing steps are provided in \autoref{Vegetation indices}. Here, we only introduce the variables and data sources of the dataset; the processing methodology is described in \autoref{Generating GEE county-level dataset for finetuning and testing}.

\subsubsection{Vegetation indices}
\label{Vegetation indices}

The Terra and Aqua combined Moderate Resolution Imaging Spectroradiometer (MODIS) Land Cover Climate Modeling Grid Version 6 product (MCD12Q1 v6) \citep{modis_vi} provides satellite-derived data at 500 m resolution and offers consistent, high-quality observations of the Earth's surface. MODIS captures information across multiple spectral bands and enables the derivation of various VIs. These indices have been widely used to monitor vegetation health, biomass, water content, and chlorophyll levels across different regions and time scales, thereby improving yield prediction by reflecting crop growth and condition. The VIs used in this study are summarized in \autoref{tab:vis} along with their equations and main applications.

\begin{table}[htbp]
  \centering
  \caption{Summary of VIs used in this study, including definitions and applications. Abbreviations: $NIR$ (near-infrared reflectance); $Red$ (red reflectance); $Blue$ (blue reflectance); $Green$ (green reflectance); $SWIR$ (shortwave infrared reflectance). Band specifications for MODIS are as follows: Red – Band 1 (620–670 nm); NIR – Band 2 (841–876 nm); Blue – Band 3 (459–479 nm); Green – Band 4 (545–565 nm); SWIR – Band 6 (1628–1652 nm).}
  \resizebox{\textwidth}{!}{
    \begin{tabular}{p{2cm} l p{5.5cm} p{6.5cm}}
    \toprule
    \textbf{Vegetation Index} & \textbf{Citation} & \textbf{Equation} & \textbf{Application} \\
    \midrule
    GCVI & \citep{gci}   & $GCVI = \frac{NIR}{Green} - 1$   & Chlorophyll content; Crop health; Nutrient status \\
    EVI  & \citep{evi}   & $EVI = \frac{2.5 \times (NIR - Red)}{NIR + 6 \times Red - 7.5 \times Blue + 1}$   & Dense canopy; High biomass; Noise reduction \\
    NDWI & \citep{ndwi}  & $NDWI = \frac{NIR - SWIR}{NIR + SWIR}$   & Soil/vegetation moisture; Drought monitoring; Irrigation management \\
    NDVI & \citep{ndvi}  & $NDVI = \frac{NIR - Red}{NIR + Red}$   & Vegetation greenness; Biomass; Yield prediction \\
    \bottomrule
    \end{tabular}
  }
  \label{tab:vis}
\end{table}

\subsubsection{Weather data}
\label{Weather data}

The Parameter-elevation Regressions on Independent Slopes Model (PRISM) dataset \citep{prism1}\citep{prism2} is a high-resolution weather dataset with 4 km resolution that provides detailed information on various climatic variables, including PPT, Tmax, and Tmin. This dataset is widely used in agricultural research, hydrology, and weather studies due to its fine spatial resolution and comprehensive coverage, making it an essential tool for understanding and predicting weather-related impacts on crop yield and other environmental processes.  

The MCD18A1 Version 6.1 \citep{modis_radn} is a MODIS Terra and Aqua combined Downward Shortwave Radiation gridded Level 3 product. The reliable radiation data is produced daily at 500 m resolution, with estimates of Downward Shortwave Radiation provided every 3 hours. Downward Shortwave Radiation is incident solar radiation over land surfaces in the shortwave spectrum (300-4,000 nanometers) and is an important variable in land-surface models that address a variety of scientific and applied issues.

\subsubsection{Soil moisture}
\label{Soil moisture} 

The SPL4SMGP.007 SMAP L4 Global 3-hourly 9-km Surface and Root Zone Soil Moisture dataset \citep{smap1, smap2} plays a critical role in our research on the relationship between drought and corn yield prediction. By providing detailed measurements of soil moisture at both the surface (0–5 cm) and root zone levels (0–100 cm), SMAP level 4 data allows us to assess the availability of water in the soil, a key factor influencing crop growth and resilience during drought conditions.

\subsubsection{Variables summary in county-level dataset for finetuning}
\label{Variables summary in county-level dataset for finetuning}

Additional features included the prediction year, location (latitude and longitude), and the 5-year historical average yield \citep{usda_nass_quickstats}. These features were also added to the GEE county-level dataset, consistent with their inclusion in the APSIM field-level dataset. All the variables in GEE county-level dataset are listed in \autoref{tab:combined_dataset}.

\section{Methodology}

\begin{figure*}[htbp]
\centering  %图片全局居中
\includegraphics[width=1\textwidth]{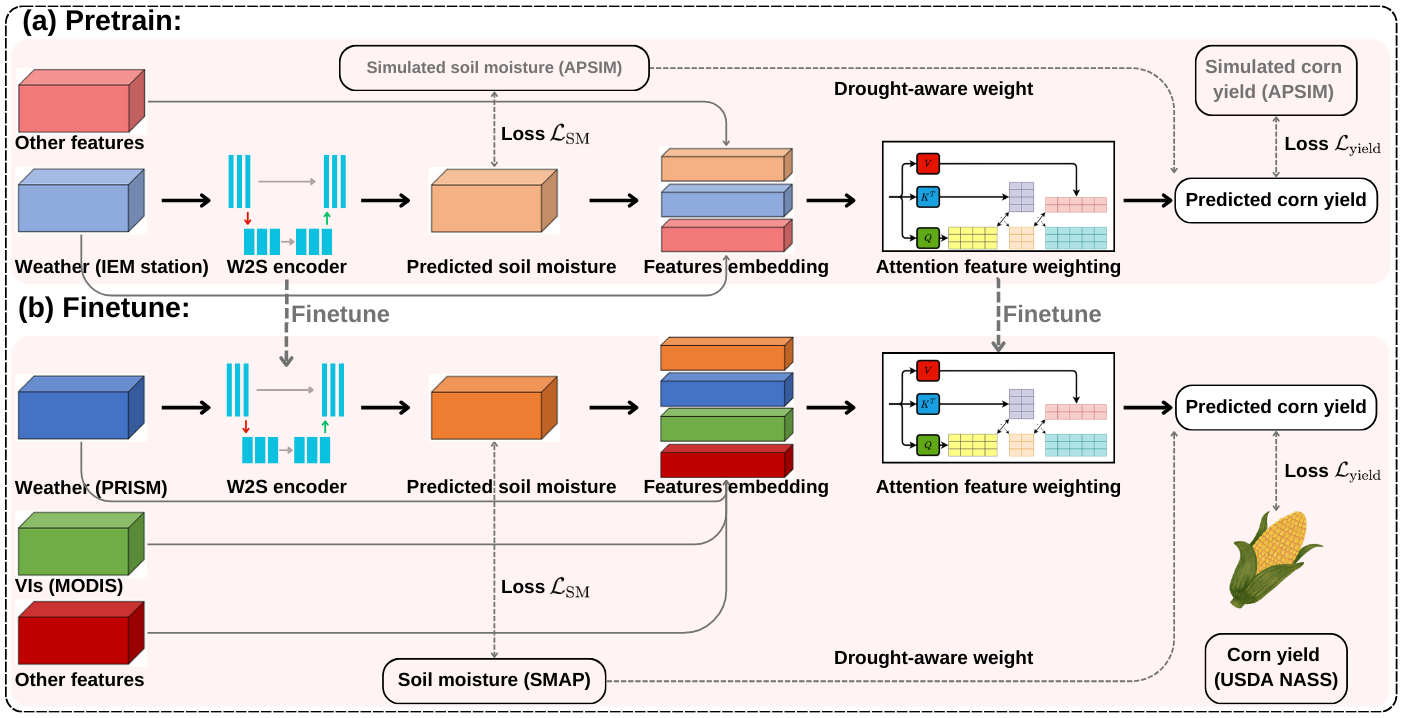}
\caption{The pipeline of the proposed KGML-SM model.}
\label{fig:pipeline}
\end{figure*}

The overall pipeline of the KGML-SM is shown in \autoref{fig:pipeline}. This architecture comprised two principal components: the W2S encoder, designed to capture the relationship between weather data and soil moisture, and the attention-based feature-weighting module, which learned how various features influence corn yield. Initially, the model was pretrained on the APSIM field-level dataset, followed by finetuning using the GEE county-level dataset.

% \clearpage

We begin with a formal problem formulation of KGML-SM (\autoref{Problem formulation}), followed by an introduction to the model components (\autoref{KGML-SM model structure}); next, we explain how the APSIM field-level dataset is generated for pretraining (\autoref{Generating APSIM field-level dataset for pretraining}) and the GEE county-level dataset for finetuning and testing (\autoref{Generating GEE county-level dataset for finetuning and testing}); then, we introduce how the KGML-SM model is trained and used for prediction (\autoref{Developing the KGML-SM framework}); after that, we describe the experimental setup (\autoref{Experimental setup}) and the statistical analysis of drought, soil moisture, and corn yield (\autoref{Statistical analysis of drought, soil moisture, and corn yield}).

\subsection{Problem formulation}
\label{Problem formulation}
% x matrix 
% soil output

The corn yield prediction problem is formally defined in this section, along with its mathematical formulation within the KGML framework. Let each unique county-year combination be represented by a sample indexed as $ i $ ($i = 1, 2, \dots, N$), where $N$ is the total number of county-year instances in the experiment. For each sample $i$, the input features are specified as follows: The temporal weather features are denoted by the vector  $ \mathbf{w_i} = [w^1_i, w^2_i, \dots, w^T_i] $. Other features, including the prediction year, geographical location, and historical average yield, are aggregated into the vector $ \mathbf{o_i} = [o^1_i, o^2_i, \dots, o^T_i] $, In these representations, $T$ corresponds to the total number of discrete time steps covering the duration from corn planting to harvest. Additionally, simulated soil moisture data is represented as $\mathbf{s_i} = [s^1_i, s^2_i, \dots, s^T_i]$. The corn yield record for sample $i$ is denoted by $y_i$ and serves as the supervision for model training.

The APSIM-generated field-level dataset, specified as $\mathcal{D}_{\text{field/pretrain}} = \{ (\mathbf{w_i}, \mathbf{o_i} \mid \mathbf{s_i}, \mathbf{y_i}) \}$, is employed for the pretraining phase. In this notation, the vertical bar $\mid$ separates the input variables ($\mathbf{w_i}, \mathbf{o_i}$) from the target labels ($\mathbf{s_i}, \mathbf{y_i}$). Furthermore, VIs represented by the vector $ \mathbf{v_i} = [v^1_i, v^2_i, \dots, v^T_i] $ are incorporated. The county-level dataset derived from GEE is denoted as $\mathcal{D}_{\text{county/finetune}} = \{ (\mathbf{w_i}, \mathbf{o_i}, \mathbf{v_i} \mid \mathbf{s_i}, \mathbf{y_i}) \}$ and is utilized for the finetuning.

The objective is to first build the W2S encoder $ f_{\text{W2S}} $ to map weather inputs to soil moisture $\mathbf{\hat{s}_i} = f_{\text{W2S}}(\mathbf{w_i})$. Then, the predicted soil moisture $ \mathbf{\hat{s}_i} $ is combined with other input features to predict yield via an attention module $ f_{\text{att}} $, resulting in the prediction $\hat{y_i} = f_{\text{att}}(\mathbf{w_i}, \mathbf{o_i}, \mathbf{v_i}, \mathbf{\hat{s}_i})
$. The model’s performance is evaluated by comparing the predicted yields $\hat{y}_i$ with the actual yields $ y_i $.

\subsection{KGML-SM model structure}
\label{KGML-SM model structure}

\subsubsection{Weather-to-Soil encoder}
\label{Weather-to-Soil encoder}

The W2S encoder is a module designed to model the influence of weather conditions on soil moisture. By capturing the statistical relationship between weather variables and soil moisture, the W2S encoder improves the representation of soil moisture dynamics at the county scale, which in turn supports more accurate yield prediction under varying weather conditions.

The W2S encoder employs a U-Net-based encoder-decoder architecture, which consists of an encoder, a decoder, and a fully connected layer \citep{dlbook} for feature transformation. Given a time-series weather input $\mathbf{w_i}$, the encoder extracts hierarchical representations by progressively downsampling the temporal dimension. The decoder then reconstructs high-level features using upsampling and skip connections that integrate information from the encoder. Finally, a fully connected layer transforms the decoded features into the predicted soil moisture output $\mathbf{\hat{s}_i}= f_{\text{W2S}}(\mathbf{w_i})$.

\subsubsection{Attention module}
\label{Attention module}

The attention mechanism \citep{attention} is a powerful tool in ML that enables models to focus on the most relevant parts of the input data when making predictions. By assigning different levels of importance to various input elements, the attention mechanism helps the model prioritize the most crucial information. In KGML-SM, we aim to use the attention mechanism to weight different features, helping us understand each feature’s contribution to yield prediction across different dimensions.

The input for corn yield prediction is formulated as $ \mathbf{X_i} =[\mathbf{w_i}; \mathbf{o_i}; \mathbf{v_i}; \mathbf{\hat{s}_i}]$ representing the concatenation of feature vectors. For each feature embedding $\mathbf{X_i}$, an attention mechanism is employed to learn the corresponding attention weight $\mathbf{\alpha_i}$. This weight is subsequently utilized in the computation of the final yield $\hat{y_i}$.

First, we compute the query $\mathbf{Q_i}$, key $\mathbf{K_i}$, and value $\mathbf{V_i}$ vectors from the feature $\mathbf{X_i}$ using learned linear transformations:

\begin{equation}
\mathbf{Q_i} = \mathbf{W_Q} \mathbf{X_i}, \quad \mathbf{K_i} = \mathbf{W_K} \mathbf{X_i}, \quad \mathbf{V_i} = \mathbf{W_V} \mathbf{X_i}
\end{equation}

where $\mathbf{W_Q}$, $\mathbf{W_K}$, and $\mathbf{W_V}$ are the learned weight matrices for the query, key, and value, respectively.

Next, we calculate the attention scores by taking the dot product of the query and key, scaled by the square root of the key's dimension $d_k$:

\begin{equation}
\text{Attention}(\mathbf{Q_i}, \mathbf{K_i}) = \frac{\mathbf{Q_i} \cdot {\mathbf{K_i}}^\top}{\sqrt{d_k}}
\end{equation}

These attention scores are then passed through a softmax function \citep{dlbook} to obtain the weights $\mathbf{\alpha_i}$:

\begin{equation}
\mathbf{\alpha_i} = \text{softmax}\left(\frac{\mathbf{Q_i} \cdot {\mathbf{K_i}}^\top}{\sqrt{d_k}}\right)
\end{equation}

The softmax function converts a vector of $K$ real numbers into a probability distribution of $K$ possible outcomes. Given a random input vector $\mathbf{z} = [z_1, \dots, z_K]$ for $ i = 1, \dots, K $, the softmax function \citep{dlbook} is defined as:

\begin{equation}
\text{softmax}(\mathbf{z})_i = \frac{e^{z_i}}{\sum_{j=1}^{K} e^{z_j}}
\end{equation}

 where $ e^{z_i} $ is the exponential of the $ i $-th element of the input vector $\mathbf{z}$.

Finally, we compute the weighted sum of the values $\mathbf{V_i}$ to obtain the final yield prediction:

\begin{equation}
\hat{y_i} = \sum_{t=1} \mathbf{\alpha_i} \mathbf{V_i}
\end{equation}

\subsection{Generating APSIM field-level dataset for pretraining}
\label{Generating APSIM field-level dataset for pretraining}

The variables comprising the field-level dataset were previously introduced (\autoref{APSIM field-level dataset}); this section provides additional processing details. Besides the input weather data obtained from IEM, specific management parameters were also necessary for model calibration (\autoref{APSIM management parameters}). Following the simulation, the resulting data required a optimizing process prior to its utilization (\autoref{Optimizing APSIM field-level dataset based on soil moisture}).

\subsubsection{APSIM management parameters}
\label{APSIM management parameters}

Simulation methodology depended on the study’s scale. For field-level areas, management information was obtained directly from farmers as input. Other parameters—such as sowing density, fertilizer amount, initial soil water content, and selected cultivar coefficients—were calibrated to optimize model performance \citep{process_4}. For county-level areas, identifying representative fields was common practice. Their management parameters were averaged, and then applied to represent the entire county \citep{process_1}. Calibrating each county individually became impractical for very large study areas like our U.S. Corn Belt simulation. In such cases, the typical approach defined a parameter range based on empirical and statistical data. These parameters were then randomly combined within specified ranges, simulating yield across all counties \citep{scym}. Given the vast area of cornfields in our study, it was not feasible to calibrate these parameters for every individual station. Rather than calibrating parameters to specific local conditions, the goal of generating simulated data was to expand the range of training samples, thereby improving the model’s capacity to generalize across diverse environments. Although this dataset may appear relatively coarse due to the lack of calibration, such diversity is in fact desirable for pretraining, as it exposes the model to a wider spectrum of conditions and enhances its ability to generalize. Importantly, our focus is not on accurately predicting yield at these station sites, but rather on ensuring robust county-level performance after finetuning with observed data. At this stage, the simulated dataset primarily serves to expose the model to general patterns of corn growth, while subsequent county-level calibration ensures predictive accuracy at the target scale. By defining parameters within broad but reasonable ranges, we were able to generate a diverse set of simulated samples, which serves as the foundation for the parameter specifications described in the following section.

Therefore, we established a general parameter range encompassing all possible values across different fields (\autoref{tab:apsim_management}). This approach ensured model applicability across a wide range of conditions, maintaining reasonable prediction accuracy. Some reasonable adjustment ranges for key management parameters were collected from prior research and USDA statistics. Sowing density determines the number of plants per unit area, directly affecting competition for resources such as light, water, and nutrients. Based on statistics from the USDA NASS \citep{usda_nass_quickstats}, the range for sowing density was set at 6-9 plants/m$^2$, values of 6, 7, 8, and 9 were used in the simulation. Sowing dates are crucial as they determine the crop’s growth cycle and its interaction with seasonal weather patterns. Typically, growers maximize corn yield by planting in late April or early May \citep{sowing_date_ia, sowing_date_mn}. In the simulation, sowing started between April 20–25 and ended between May 15–20. Fertilizer application is essential for providing the necessary nutrients to support plant growth. The most commonly used nitrogen fertilizers for corn production in North America are anhydrous ammonia, urea, and urea-ammonium nitrate solutions \citep{iowaiem}. The fertilizer amount was set at 200-300 kg/ha of urea nitrogen (N). Values of 200, 250, and 300 were used in the simulation. Initial soil water content is important for establishing the starting conditions for the model’s simulation of soil moisture dynamics throughout the growing season. The initial soil water content was set between 40\% and 60\%, values of 40\%, 50\%, and 60\% were used in the simulation.

\begin{table}[htbp]
  \centering
  \caption{Summary of management parameters in APSIM simulation}
  \scalebox{0.8}{
    \begin{tabular}{lll}
    \toprule
    Factor & \multicolumn{1}{l}{Value range} & \multicolumn{1}{l}{Source} \\
    \midrule
    Start of sowing window &   Apr-20 to Apr-25    & \citep{sowing_date_ia, sowing_date_mn} \\
    End of sowing window &   May-15 to May-20    & \citep{sowing_date_ia, sowing_date_mn, scym} \\
    Plant population &    6-9 plants/m$^2$   & \citep{usda_nass_quickstats} \\
    Fertilizer amount &    200-300 kg/ha   & \citep{iowaiem} \\
    Intial soil water &   40\%-60\%    & \citep{scym} \\
    \bottomrule
    \end{tabular}%
    }
  \label{tab:apsim_management}%
\end{table}%
% \clearpage

\subsubsection{Optimizing APSIM field-level dataset based on soil moisture}
\label{Optimizing APSIM field-level dataset based on soil moisture}

APSIM was not accurate for large-scale corn yield simulation without precise management adjustments, making it necessary to filter the APSIM field-level dataset before using it for pretraining. Although the goal was to construct a diverse dataset, retaining highly unrealistic samples would only introduce noise and bias the model away from biologically plausible relationships. Filtering was therefore essential to ensure that the dataset remained both diverse and reliable.

Soil moisture was selected as the benchmark variable because it provides a balance between data availability and process relevance. Although it is influenced by multiple factors such as PPT, soil texture, and topography, detailed station-level information on these factors was unavailable across the study region. In contrast, weather data were consistently available and strongly correlated with soil moisture, making it a practical proxy for dataset screening. Moreover, cumulative simulation errors are typically smaller for soil moisture than for yield, which is affected by many interacting processes. Based on these considerations, we used soil moisture quality as the criterion to optimize the APSIM field-level dataset.

Specifically, a Linear Regression (LR) model was trained on the GEE county-level dataset to predict soil moisture from four weather inputs (Tmax, Tmin, PPT, Radn). The trained model then predicted soil moisture predictions for the APSIM field-level dataset, which were compared against the APSIM-simulated values. To ensure data quality while retaining sufficient variability, samples with mean squared error (MSE) above 0.5 were discarded. This threshold was chosen as a balance: higher-error samples (e.g., $\geq$1) were unreliable compared to typical soil moisture RMSE values, yet a more stringent cutoff could have removed too much variability and reduced the generalization capacity of the pretraining dataset.

\subsection{Generating GEE county-level dataset for finetuning and testing}
\label{Generating GEE county-level dataset for finetuning and testing}

\subsubsection{Feature extraction within corn field}
\label{Feature extraction within corn field}

To focus on the corn portion of the acquired feature data from GEE, the cornfield areas had to be identified. The Cropland Data Layer (CDL) cropland mask \citep{cdl} was used to extract the corn class for each year from 2015 to 2023. The CDL is an annual raster-based dataset with a 30-meter resolution, providing crop-specific land cover information produced by the USDA. These annual cornfield masks were employed in GEE to support our corn yield prediction study in twelve U.S. Corn Belt states.

\subsubsection{Data preprocessing}
\label{Data preprocessing}

For each county, pixel-level feature values from MODIS and SMAP products were first averaged spatially to obtain county-level values. These county-level values were then aggregated into a single representative mean value every 16 days during the growing season (April–October), following the experimental setup of previous research \citep{yuchi0, att_model}. This temporal aggregation balanced data availability with noise reduction and provided a consistent seasonal trajectory of vegetation and environmental conditions for each county. We noted that MODIS standard products included atmospheric correction and provided quality assurance layers that flag cloud-affected or low-quality pixels. In this study, we did not apply additional cloud filtering beyond the standard MODIS product processing; instead, the 16-day temporal aggregation helps to smooth residual noise caused by occasional cloud contamination. This procedure resulted in an annual time-series vector of features for each county, which served as input for model training and evaluation.

\subsection{Developing the KGML-SM framework}
\label{Developing the KGML-SM framework}

\subsubsection{Pretraining with APSIM field-level dataset}
\label{Pretraining with APSIM field-level dataset}

The pretraining process began by using a W2S encoder $f_{W2S}$ to learn the relationship between input weather features $\mathbf{w_i}$ and predicted soil moisture $\mathbf{\hat{s}_i}$. This process was formulated as $ \mathbf{\hat{s}_i} = f_{W2S}(\mathbf{w_i}),\ \text{where} \ \mathbf{w_i} \in \mathcal{D}_{\text{field/pretrain}} $.  Then, The simulated soil moisture $\mathbf{s_i} \in \mathcal{D}_{\text{field/pretrain}} $ is used to guide the W2S encoder. The loss function $\mathcal{L}_{\text{SM}}$ was defined as:

\begin{equation}
\mathcal{L}_{\text{SM}} = \frac{1}{N} \sum_{i=1}^N (\mathbf{s_i} - \mathbf{\hat{s}_i})^2
\end{equation}

Next, the predicted soil moisture was concatenated with weather data and other features to form the features embedding $\mathbf{X_i} = [\mathbf{w_i}; \mathbf{o_i}; \mathbf{\hat{s}_i}]$. The final corn yield $\hat{y}_i$ was predicted using the attention module $ f_{att} $, with the concatenated features $\mathbf{X_i}$ as input:

\begin{equation}
\hat{y}_i = f_{att}(\mathbf{X_i})
\end{equation}

\subsubsection{Drought-aware yield prediction loss function}
\label{Drought-aware yield prediction loss function}

Soil moisture is widely recognized to affect crop yield \citep{sm_research_1, sm_research_2} and numerous studies explored this relationship \citep{sm_research_3, sm_research_4, sm_research_5}. However most prior studies lacked an elegant quantitative approach addressing this issue. This study introduces a loss function adjusting predicted corn yield based on varying soil moisture levels. The final objective function jointly optimizes both components and is defined as:
\begin{equation}
\mathcal{L}_{total} = \mathcal{L}_{SM} + \mathcal{L}_{yield}
\end{equation}

The yield prediction loss function was designed to improve the model’s accuracy while incorporating drought sensitivity and penalizing overestimation. It was formulated as:

\begin{equation}
\mathcal{L}_{\text{yield}} = \frac{1}{N} \sum_{i=1}^{N} d_i 
\left[ (y_i - \hat{y}_i)^2 + \lambda \max(0, \hat{y}_i - y_i)^2 \right]
\end{equation}

where $ d_i $ is a drought-aware weighting factor, which was defined as:

\begin{equation}
d_i = \frac{1}{\bar{s}_i + \varepsilon}
\end{equation}

where $ \bar{s}_i $ is the average soil moisture during the growing season (April–October) for sample $ i $, calculated from the input soil moisture dataset (APSIM-simulated data for pretraining, and GEE/SMAP soil moisture for finetuning). $ \varepsilon $ is a small constant to prevent numerical instability, which we set to 1 in our experiment. Since soil moisture plays a critical role in crop growth and yield formation, the loss function assigned a higher penalty to drier conditions, encouraging the model to be more responsive to soil moisture variations.

Additionally, there is an asymmetric penalty term \citep{asymmetric} controlled by the factor $ \lambda > 0 $, which amplifies the loss when the predicted yield $ \hat{y}_i $ exceeds the true yield $ y_i $:

\begin{equation}
\lambda \max(0, \hat{y}_i - y_i)^2
\end{equation}

This asymmetry discourages overestimation, particularly under drought conditions, where yield predictions tend to be more uncertain. By applying a stronger penalty to overestimated yields, the model is encouraged to be more conservative, reducing the risk of unrealistic predictions. The value of hyperparameter $ \lambda $ was selected based on validation performance, and we set $ \lambda = 2 $ in our experiment (\autoref{Ablation study of different compoments in KGML-SM model}).

Specifically, when $\hat{y}_i > y_i$, the predicted error is scaled by a penalty factor $\lambda$, amplifying the loss in these cases. This encourages the model to adopt a conservative approach, reducing the likelihood of overestimating yield, particularly in drought-prone regions where overestimation could lead to inaccurate agricultural planning.

This whole loss function formulation ensured that the model not only learned accurate yield predictions but also captured the impact of soil moisture variability and drought stress, leading to more reliable and interpretable results.

\subsubsection{Finetuning and testing with GEE county-level dataset}
\label{Finetuning and testing with GEE county-level dataset}

Following the pretraining of the model on the APSIM field-level dataset $\mathcal{D}_{\text{field/pretrain}}$, finetuning using the GEE county-level dataset $\mathcal{D}_{\text{county/finetune}}$ was required to improve the model for the county-level corn yield prediction task. This process involved initially partitioning the data into training, validation, and test sets, followed by predicting corn yield for the target years. We ensured independent validation by adopting a temporal split strategy, where models were trained on preceding years and tested on the target year. Specific details regarding the partition and utilization of these datasets are elaborated upon in the experimental setup (\autoref{Experimental setup}). The loss functions $\mathcal{L}_\text{yield}$ and $\mathcal{L}_\text{SM}$ were also optimized on the training and validation datasets.

\subsection{Experimental setup}
\label{Experimental setup}

We conducted experiments on both traditional ML models and DL models. When predicting corn yield for a specific year, we trained the model using all data from preceding years, then split the dataset into 80\% for training and 20\% for validation, and tested it on the target year. Each experiment was conducted five times with different random seeds, and the final results represented the average across these runs to ensure robustness and reliability.

We implemented the DL models using the PyTorch framework \citep{pytorch} and the traditional ML code with sklearn \citep{sklearn}. The models were run on A100-SXM4-40GB and A100-SXM4-80GB GPUs. For pretraining, we used a batch size of 64, a learning rate of 0.001, the Adam optimizer \citep{adam}, and the ReduceLROnPlateau scheduler with a patience of 5, with training stopped once the RMSE dropped below 1. For finetuning, we used a smaller batch size of 16 but kept the same learning rate, optimizer, and scheduler, and applied early stopping based on validation loss. The full configurations are summarized in \autoref{tab:train_config}.

\begin{table}[htbp]
  \centering
  \caption{Training configurations for pretraining and finetuning.}
  \scalebox{0.65}{
    \begin{tabular}{lllllll p{4cm}}
    \toprule
    Stage & \makecell{Batch \\ size} &  \makecell{Learning \\ rate}  & Optimizer & Scheduler & Epochs & Loss function & Stopping criterion \\
    \midrule
    Pretrain & 64 & 0.001 & Adam & \makecell{ReduceLROnPlateau \\ (patience=5)}   & $\sim$50 & MSE  & Training RMSE $<$ 1 \\
    Finetune  & 16 & 0.001 & Adam & \makecell{ReduceLROnPlateau \\ (patience=5)} & $\sim$30 & MSE & Early stopping based on validation loss \\
    \bottomrule
    \end{tabular}
  }
  \label{tab:train_config}
\end{table}

Root mean square error ($RMSE$) and the coefficient of determination ($R^2$) were used to evaluate the performance of our model. The formulas for $RMSE$ and $R^2$ are: 

\begin{equation}
    RMSE = \sqrt{\frac{1}{n} \sum_{i=1}^{n} (y_i - \hat{y}_i)^2}
\end{equation}

\begin{equation}
    R^2 = 1 - \frac{\sum_{i=1}^{n} (y_i - \hat{y}_i)^2}{\sum_{i=1}^{n} (y_i - \bar{y})^2}
\end{equation}

where $n$ is the number of observations, $y_i$ is the actual value for the $i$-th observation, $\hat{y}_i$ is the predicted value for the $i$-th observation, and $\bar{y}$ is the mean of the actual values.

\subsection{Statistical analysis of drought, soil moisture, and corn yield}
\label{Statistical analysis of drought, soil moisture, and corn yield}

To study the impact of soil moisture on corn yield in the KGML-SM model and provide interpretability, we first analyzed the statistics of drought, soil moisture, and corn yield. The analysis aims to determine which regions experienced drought and reduced corn yield, and their relationship with soil moisture.

\subsubsection{Specifying drought area}
\label{Specifying drought area}

\begin{figure*}[htbp]
\centering  
\includegraphics[width=1\textwidth]{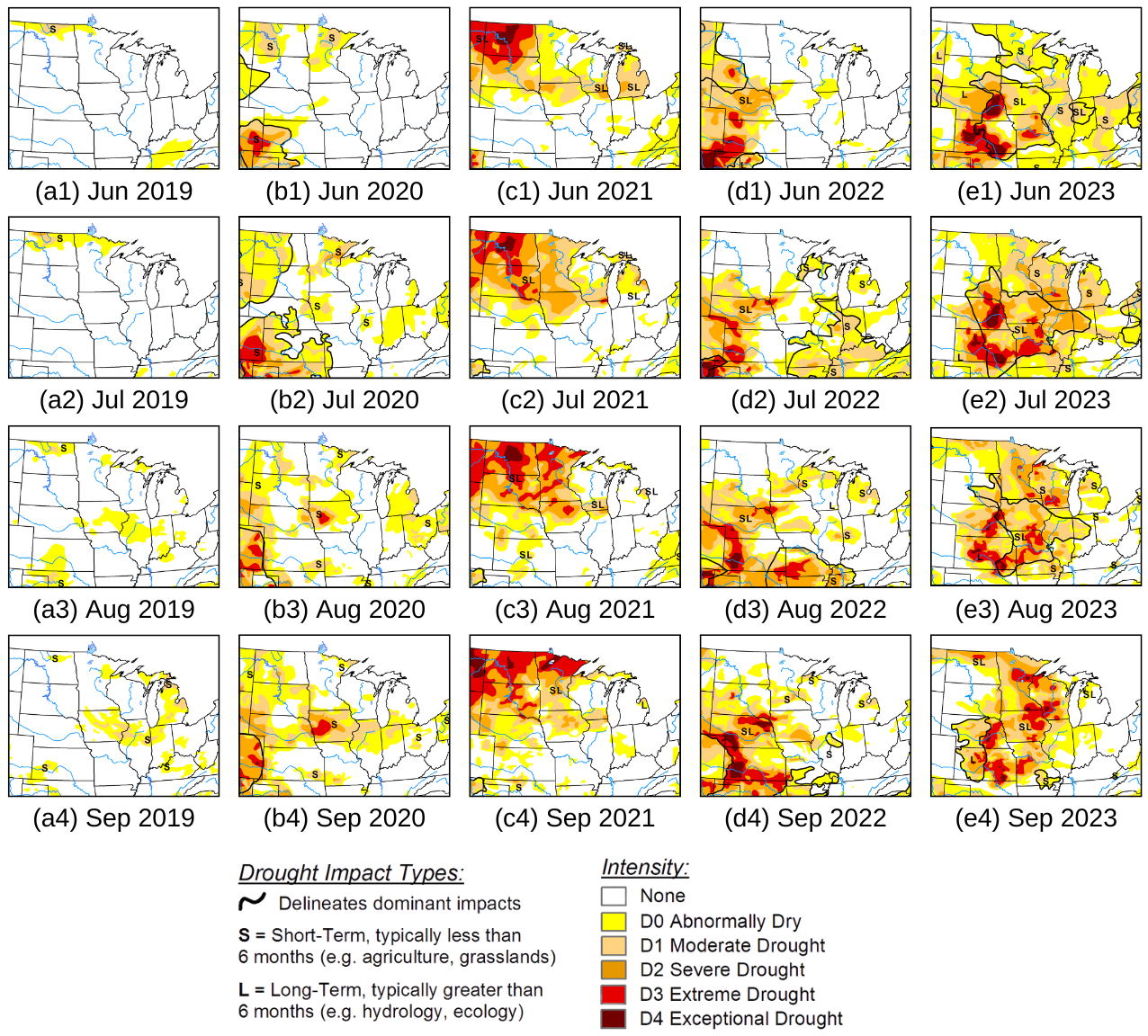}
\caption{Drought map of the U.S. for (1) June -(4) September from (a) 2019 to (e) 2023.}
\label{fig:droughtmap}
\end{figure*}

To objectively determine drought conditions, we used county‐level data from the U.S. Drought Monitor (USDM) \citep{drought_web}. For each Corn Belt state during the corn growing season (June–September; \citep{corn_harvest}), the USDM provides categorical drought classifications ranging from None to D4 (Exceptional Drought). \autoref{fig:droughtmap} shows the spatial distribution of these drought categories, providing a visual representation of drought intensity across years. To quantitatively assess drought severity, we further aggregated the USDM records of county areas under each drought category (None–D4). By summing the areas across drought categories, we calculated the proportion of each state’s total area that experienced drought in a given year. To complement the maps,  \autoref{tab:drought_stat} reports the state‐level proportions of drought‐affected counties, with values exceeding 30\% highlighted in bold to emphasize years and regions with particularly widespread drought.

\clearpage

\begin{table}[htbp]
\centering
\caption{Proportion of counties affected by drought (categories D0–D4) in each of the 12 U.S. Corn Belt states (2018–2023), derived from U.S. Drought Monitor (USDM) county-level records during the corn growing season (June–September). Values indicate the fraction of state area under drought, with those exceeding 30\% highlighted in bold.}
\label{tab:drought_pct}
\resizebox{\textwidth}{!}{%
\begin{tabular}{c|cccccccccccc}
\hline
Year & IA & IL & IN & KS & MI & MN & MO & ND & NE & OH & SD & WI \\
\hline
% 2018 & \textbf{36\%} & 25\% & 12\%   & \textbf{86\%} & \textbf{45\%} & 27\% & \textbf{85\%} & \textbf{59\%} & 11\% & 6\%  & \textbf{40\%} & 24\% \\
2019 & 17\% & 16\% & 15\%   & 9\%  & 15\% & 11\% & 1\%  & \textbf{37\%} & 1\%  & 4\%  & 0\%  & 1\%  \\
2020 & \textbf{64\%}& 19\% & \textbf{49\%}   & \textbf{64\%} & 22\% & \textbf{48\%} & 19\% & \textbf{68\%} & \textbf{55\%} & \textbf{47\%} & \textbf{59\%} & 8\%  \\
2021 & \textbf{89\%} & 25\% & 18\%   & 30\% & \textbf{65\%} & \textbf{100\%}& 8\%  & \textbf{100\%}& \textbf{76\%} & 12\% & \textbf{100\%}& \textbf{66\%} \\
2022 & \textbf{62\%} & \textbf{38\%} & \textbf{47\%}   & \textbf{88\%} & 30\% & 23\% & \textbf{61\%} & 8\%  & \textbf{97\%} & 9\%  & \textbf{73\%} & \textbf{37\%} \\
2023 & \textbf{100\%}& \textbf{94\%} & \textbf{73\%}   & \textbf{97\%} & \textbf{72\%} & \textbf{96\%} & \textbf{95\%} & \textbf{57\%} & \textbf{93\%} & \textbf{58\%} & \textbf{61\%} & \textbf{98\%} \\
\hline
\end{tabular}%
}
\label{tab:drought_stat}
\end{table}

\subsubsection{Soil moisture statistics}
\label{Soil moisture statistics}

\begin{figure*}[htbp]
\centering  %图片全局居中
\includegraphics[width=1\textwidth]{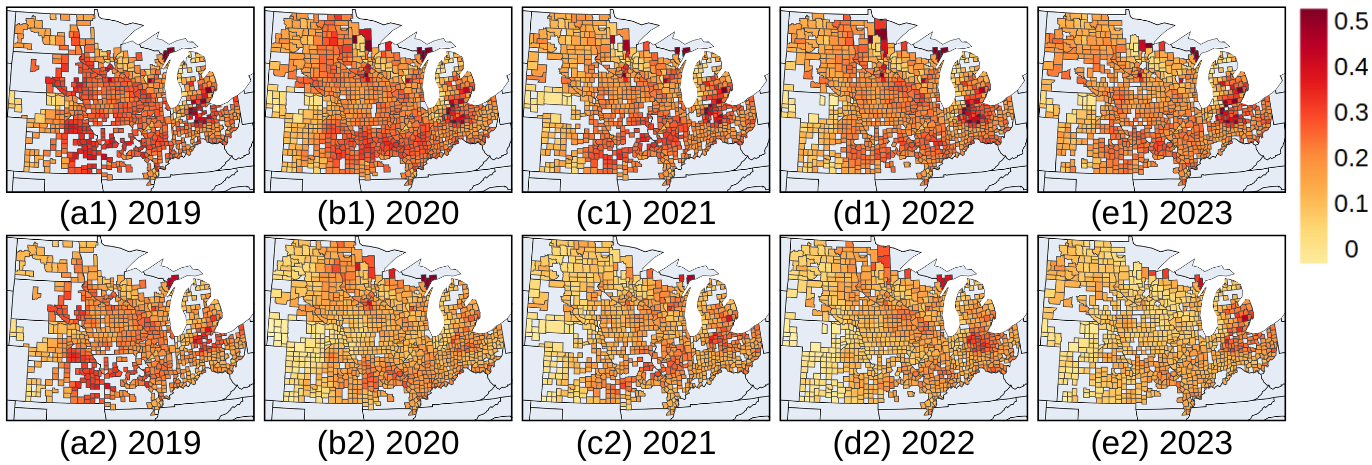}
\caption{This figure shows the maps of average (1) rootzone  and (2) surface  soil moisture during June-September from  (a) 2019  to (e) 2023.}
\label{fig:sm_map}
\end{figure*}

Next, we specified the relationship between soil moisture and drought. \autoref{fig:sm_map} shows the maps of average rootzone and surface soil moisture during June-September from 2019 to 2023. The comparison of the drought maps above with soil moisture data from 2019 to 2023 revealed a strong correlation between drought-affected areas and lower soil moisture levels. We also noticed that rootzone moisture was more abundant than surface moisture and that the two spatial distributions were generally similar.

% \clearpage

\subsubsection{Drought impact on corn yield}
\label{Drought impact on corn yield}

\begin{figure*}[htbp]
\centering  %图片全局居中
\includegraphics[width=1\textwidth]{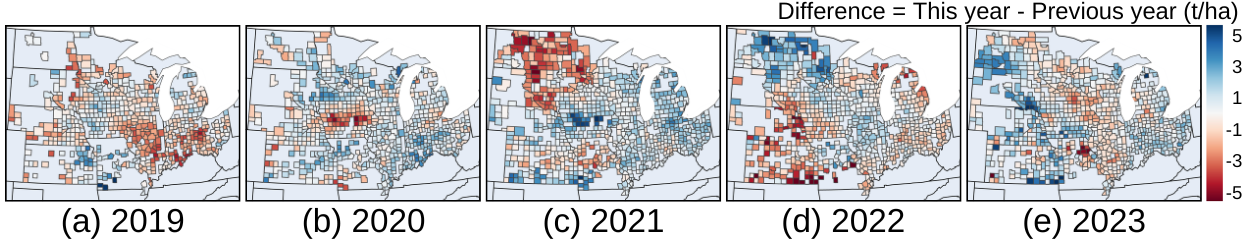}
\caption{The difference in corn yield between each year from (a) 2019 to (e) 2023 and the previous year, with negative values indicating a reduction in yield.}
\label{fig:this_vs_previous_yield}
\end{figure*}

Finally, we examined the interannual impact of drought on corn yield.  \autoref{fig:this_vs_previous_yield} shows the county-level yield differences between each year from 2019 to 2023 and the previous year, where negative values indicate yield reductions. To further quantify the extent of yield losses, we calculated for each state the proportion of counties experiencing a yield decline greater than 1 t/ha relative to the previous year. These results are summarized in \autoref{tab:reduced_yield_stat}, with values above 30\% highlighted in bold to emphasize states and years with widespread yield reductions.

\begin{table}[htbp]
\centering
\caption{Proportion of counties with yield reductions greater than 1 t/ha relative to the previous year across 12 Corn Belt states (2019–2023), with those exceeding 30\% highlighted in bold.}
\resizebox{\textwidth}{!}{%
\begin{tabular}{c|cccccccccccc}
\hline
Year & IA & IL & IN & KS & MI & MN & MO & ND & NE & OH & SD & WI \\
\hline
2019 & 27\% & \textbf{78\%} & \textbf{67\%} & 15\% & 21\% & \textbf{36\%} & 10\% & \textbf{50\%} & \textbf{32\%} & \textbf{54\%} & \textbf{42\%} & \textbf{30\%} \\
2020 & \textbf{54\%} & 1\%  & 2\%  & 20\% & 8\%  & 5\%  & 17\% & \textbf{33\%} & 16\% & 7\%  & 4\%  & 7\%  \\
2021 & 1\%  & 0\%  & 4\%  & 16\% & 0\%  & \textbf{50\%} & \textbf{42\%} & \textbf{87\%} & 3\%  & 0\%  & \textbf{71\%} & 0\%  \\
2022 & \textbf{33\%} & 1\%  & \textbf{30\%} & \textbf{76\%} & \textbf{42\%} & 3\%  & 28\% & 0\%  & \textbf{85\%} & 26\% & \textbf{39\%} & 12\% \\
2023 & 21\% & 23\% & 1\%  & 15\% & 11\% & \textbf{30\%} & \textbf{36\%} & 17\% & 8\%  & 0\%  & 2\%  & 17\% \\
\hline
\end{tabular}%
}
\label{tab:reduced_yield_stat}
\end{table}

\subsubsection{Statistical summary of drought-affected states}
\label{Statistical summary of drought-affected states}

Based on the above statistical analysis, we summarized the drought-affected areas that experienced yield reductions in \autoref{tab:drought_reduced_yield_stat}. This table shows the states where drought and yield loss overlapped. In 2019, drought was limited, but many counties still had large yield reductions. This was mainly due to delayed planting and record prevented planting caused by heavy rainfall and flooding \citep{fb2019preventplant, usda2019preventplant}. In 2020, fewer yield losses were observed, likely because many states had already lost yield in 2019, leaving less room for further decline. In 2021 and 2022, drought-affected states also showed clear yield reductions, which matches the expected pattern. In 2023, drought was widespread across all states, but only a few states had obvious yield losses, because severe drought in 2022 had already caused large yield reductions.

% \clearpage

\begin{table}[htbp]
  \centering
  \caption{Summary of obviously yield-reduced states within drought-affected areas.}
  \scalebox{0.7}{
    \begin{tabular}{clll}
    \toprule
    \diagbox{Year}{Area type} & Drought-affected states & Obviously yield-reduced states & Intersection \\
    \midrule
    % 2018 & KS,MO,ND,MI,SD,IA & - &  \\
    2019 & ND & IL,IN,MN,ND,NE,OH,SD,WI  & ND \\
    2020 & ND,IA,KS,SD,NE,MN,IN,OH & IA,ND & IA,ND \\
    2021 & MN,ND,SD,IA,NE,WI,MI & MN,MO,ND,SD & MN,ND,SD \\
    2022 & NE,KS,SD,IA,MO,IN,IL,WI & IA,IN,KS,MI,NE,SD  & IA,IN,KS,NE,SD\\
    2023 & All the states & MN,MO  & MN,MO\\
    \bottomrule
    \end{tabular}%
    }
  \label{tab:drought_reduced_yield_stat}%
\end{table}%

\section{Experimental Results}
\label{Experimental Results}

% \clearpage

\subsection{Evaluation results}
\label{Evaluation results}

To validate the superiority of our KGML-SM model, we compared it with some commonly used ML models in remote sensing: LR, multilayer perceptron (MLP), ridge regression (RR), and random forest (RF). LR \citep{lr} is a simple statistical method that models the relationship between a dependent variable and one or more independent variables by fitting a linear equation to the data. MLP \citep{dlbook} is a neural network with multiple layers, including an input, hidden, and output layer. It captures non-linear relationships and is widely used for classification and regression. RR \citep{ridge} is an extension of linear regression that includes an L2 regularization term to prevent overfitting by penalizing large coefficients. RF \citep{rf} is an ensemble learning method that constructs multiple decision trees during training and aggregates their predictions to improve accuracy and robustness.

The results showed that our KGML-SM model consistently outperformed other ML models across all years. As shown in \autoref{tab:result_tradition_model}, RF performed the best among traditional ML models, indicating its strong ability to capture complex relationships in the data \citep{rf_support}. RR performed slightly worse than RF, with slightly higher RMSE values, suggesting that regularization helped improve predictions but was not as effective as ensemble learning \citep{ridge_support}. LR ranked next, showing higher RMSE values, likely due to its inability to model non-linear relationships effectively \citep{lr_support}. MLP performed the worst, with the highest RMSE values in most years, indicating that it struggled to generalize well, possibly due to overfitting or insufficient training data \citep{mlp_support}.

\begin{table}[htbp]
  \centering
  \caption{Comparison with traditional ML models}
  \scalebox{0.70}{
    \begin{tabular}{rrrrrrrrrrr}
    \toprule
    \diagbox{\textbf{Year}}{\textbf{Method}} & \multicolumn{2}{r}{KGML-SM} & \multicolumn{2}{r}{LR} & \multicolumn{2}{r}{MLP} & \multicolumn{2}{r}{RR} & \multicolumn{2}{r}{RF} \\
\cmidrule{2-11}          & RMSE  & R2    & RMSE  & R2    & RMSE  & R2    & RMSE  & R2    & RMSE  & R2 \\
    \midrule
    2019  & \textbf{0.964} & \textbf{0.741} & 1.328 & 0.621 & 1.169 & 0.607 & 1.214 & 0.607 & 1.040  & 0.712 \\
    2020  & \textbf{0.980} & \textbf{0.792} & 1.304 & 0.690  & 1.207 & 0.734 & 1.230  & 0.661 & 1.120  & 0.719 \\
    2021  & \textbf{1.104} & \textbf{0.836} & 1.167 & 0.790  & 1.247 & 0.761 & 1.129 & 0.808 & 1.236 & 0.794 \\
    2022  & \textbf{1.085} & \textbf{0.837} & 1.471 & 0.740  & 1.400   & 0.765 & 1.318 & 0.781 & 1.185 & 0.821 \\
    2023  & \textbf{1.071} & \textbf{0.807} & 1.226 & 0.737 & 1.225 & 0.738 & 1.140  & 0.776 & 1.196 & 0.791 \\
    \bottomrule
    \end{tabular}%
    }
  \label{tab:result_tradition_model}%
\end{table}%

% \clearpage

\subsection{Ablation study of different components in KGML-SM model}
\label{Ablation study of different compoments in KGML-SM model}

To further demonstrate the contribution of each module in our KGML-SM model, we conducted a series of ablation studies. We began with an attention-based baseline model without soil moisture inputs (Att w/o SM), which used all features except soil moisture. We then included soil moisture in the input features to form the Att model. Next, we incorporated the APSIM field-level dataset for pretraining (Att+sim), followed by the addition of the W2S encoder to integrate soil moisture dynamics (Att+sim+W2S). To further analyze the contribution of different loss components in our drought-aware loss $\mathcal{L}_{\text{yield}} = \frac{1}{N} \sum_{i=1}^{N} d_i \left[ (y_i - \hat{y}_i)^2 + \lambda \max(0, \hat{y}_i - y_i)^2 \right]$, we first added the Soil Moisture Weighted (SMW) term, represented by $d_i$, which assigns greater weight to errors under low soil moisture conditions (Att+sim+W2S+SMW). We then introduced the Overestimation penalty (OE) term, $\lambda \max(0, \hat{y}_i - y_i)^2$, which penalizes yield overestimation more strongly, particularly under drought conditions. Combining both SMW and OE yielded the full KGML-SM framework. A summary of the components of each ablation model is provided in \autoref{tab:ablation_methods}.

\begin{table}[htbp]
  \centering
  \caption{Comparison of different methods and their components.}
  \scalebox{0.7}{
    \begin{tabular}{lcccccc}
    \toprule
    Method & \multicolumn{1}{l}{\makecell{Attention\\ module}}& \multicolumn{1}{l}{\makecell{Soil\\ moisture}}  & \multicolumn{1}{l}{\makecell{Field-level data\\ pretraining}} & \multicolumn{1}{l}{\makecell{W2S\\ encoder}} & \multicolumn{1}{l}{\makecell{SMW\\ loss}} & \multicolumn{1}{l}{\makecell{OE \\ loss}} \\
    \midrule
    Att w/o SM   & \checkmark  &     &       &       &  \\
    Att   & \checkmark  &   \checkmark    &       &       &  \\
    Att+sim & \checkmark  &    \checkmark   &    \checkmark   &       &  & \\
    Att+sim+W2S &  \checkmark  &  \checkmark & \checkmark & \checkmark & & \\
    Att+sim+W2S+SMW & \checkmark  &   \checkmark    &    \checkmark   &   \checkmark    & \checkmark & \\
    KGML-SM & \checkmark   &   \checkmark   &  \checkmark     &     \checkmark  & \checkmark  & \checkmark \\
    \bottomrule
    \end{tabular}%
    }
  \label{tab:ablation_methods}%
\end{table}%

% \clearpage

Through the ablation study of all model components (\autoref{tab:ablation_result}), we found that the APSIM field-level dataset pretraining contributed the most to performance improvement. This indicated that our APSIM field-level dataset effectively captured county-level data patterns, playing a crucial role in enabling the model to learn the relationship between corn yield and agricultural variables. When comparing the baseline attention model without soil moisture (Att w/o SM) to the one including soil moisture inputs (Att), the performance difference was marginal across most years. This result suggests that simply adding raw soil moisture values provides limited benefits, highlighting the necessity of our subsequent modules to better exploit soil moisture information.

Additionally, our drought-aware components also contributed to performance gains. While the SMW Loss improved performance in some drought-affected years, the effect was not consistent. This is because overestimation of yield can arise not only from drought but also from other natural hazards such as flooding, extreme temperatures, or pest outbreaks, which are not fully captured by soil moisture weighting alone. In contrast, the overestimation penalty directly constrains the model against overprediction across diverse adverse conditions, thereby providing more robust improvements.

% \clearpage
% Transposed ablation performance table with RMSE and R2 under each year
\begin{table}[htbp]
  \centering
  \caption{Ablation study of different components in KGML-SM model (RMSE and R$^2$ across years).}
  \scalebox{0.7}{
    \begin{tabular}{lcccccccccc}
    \toprule
    % \textbf{Method} 
    \diagbox{\textbf{Method}}{\textbf{Year}}
      & \multicolumn{2}{c}{2019} 
      & \multicolumn{2}{c}{2020} 
      & \multicolumn{2}{c}{2021} 
      & \multicolumn{2}{c}{2022} 
      & \multicolumn{2}{c}{2023} \\
    \cmidrule(lr){2-3}\cmidrule(lr){4-5}\cmidrule(lr){6-7}\cmidrule(lr){8-9}\cmidrule(lr){10-11}
      & RMSE & R$^2$ 
      & RMSE & R$^2$ 
      & RMSE & R$^2$ 
      & RMSE & R$^2$ 
      & RMSE & R$^2$ \\
    \midrule
    Att w/o SM       & 1.258  & 0.578  & 1.053  & 0.752  & 1.149  & 0.812  & 1.315  & 0.780  &  1.096 &  0.784 \\
    
    Att              & 1.268 & 0.570 & 1.011 & 0.770 & 1.195 & 0.808 & 1.315 & 0.779 & 1.201 & 0.766 \\
    
    Att+sim          & 1.087 & 0.715 & 1.003 & 0.770 & 1.143 & 0.814 & 1.144 & 0.802 & 1.114 & 0.781 \\
    
    Att+sim+W2S      & 0.974 & 0.732 & 0.981 & 0.783 & 1.127 & 0.811 & 1.119 & 0.802 & 1.101 & 0.805 \\
    
    Att+sim+W2S+SMW   & 1.097  & 0.711  & 1.054  & 0.763  & 1.112  & 0.832  & \textbf{1.074}  &  0.821 & \textbf{1.043}  & \textbf{0.812}  \\
    
    KGML-SM          & \textbf{0.964} & \textbf{0.741} & \textbf{0.980} & \textbf{0.792} & \textbf{1.104} & \textbf{0.836} & 1.085 & \textbf{0.837} & 1.071 & 0.807 \\
    \bottomrule
    \end{tabular}
  }
  \label{tab:ablation_result}
\end{table}

In DL, loss functions often include hyperparameters that control the relative importance of different error components \citep{dlbook}. In the drought-aware loss, the coefficient $\lambda$ determines the penalty strength applied to yield overestimation. Choosing an appropriate value of $\lambda$ is therefore critical: too small a value would fail to constrain overestimation effectively, whereas too large a value could distort the optimization and harm overall accuracy. To ensure a principled selection, we performed hyperparameter tuning on the validation set by testing multiple candidate values ($\lambda=0,1,2,5,10$) rather than assigning it arbitrarily. The results are summarized in \autoref{tab:lambda_ablation}. We found that $\lambda=2$ consistently offered the best trade-off between reducing RMSE and improving $R^2$, while both smaller and larger values led to inferior performance. Consequently, we adopted $\lambda=2$ as the default setting in the KGML-SM framework.

\begin{table}[htbp]
  \centering
  \caption{Ablation study on different values of the overestimation penalty coefficient $\lambda$ (RMSE and R$^2$ for 2019–2023). The best $\lambda$ was selected based on validation performance.}
  \scalebox{0.8}{
    \begin{tabular}{lcccccccccc}
    \toprule
    \multirow{2}{*}{\textbf{Year}} 
      & \multicolumn{2}{c}{$\lambda=0$} 
      & \multicolumn{2}{c}{$\lambda=1$} 
      & \multicolumn{2}{c}{$\lambda=2$} 
      & \multicolumn{2}{c}{$\lambda=5$} 
      & \multicolumn{2}{c}{$\lambda=10$} \\
    \cmidrule(lr){2-3}\cmidrule(lr){4-5}\cmidrule(lr){6-7}\cmidrule(lr){8-9}\cmidrule(lr){10-11}
      & RMSE & R$^2$ 
      & RMSE & R$^2$ 
      & RMSE & R$^2$ 
      & RMSE & R$^2$ 
      & RMSE & R$^2$ \\
    \midrule
    2019 & 1.097 & 0.711 & 0.982 & \textbf{0.752} & \textbf{0.964} & 0.741 & 1.032 & 0.730 & 1.115 & 0.695 \\
    
    2020 & 1.054 & 0.763 & \textbf{0.976} & 0.775 & 0.980 & \textbf{0.792} & 1.021 & 0.749 & 1.109 & 0.710 \\
    
    2021 & 1.112 & 0.832 & 1.122 & 0.819 & \textbf{1.104} & \textbf{0.836} & 1.166 & 0.785 & 1.235 & 0.745 \\
    
    2022 & \textbf{1.074} & 0.821 & 1.108 & 0.829 & 1.085 & \textbf{0.837} & 1.153 & 0.792 & 1.223 & 0.755 \\
    
    2023 & \textbf{1.043} & \textbf{0.812} & 1.062 & 0.801 & 1.071 & 0.807 & 1.184 & 0.776 & 1.246 & 0.740 \\
    \bottomrule
    \end{tabular}
  }
  \label{tab:lambda_ablation}
\end{table}

To evaluate the effect of filtering the simulated dataset (\autoref{Optimizing APSIM field-level dataset based on soil moisture}), we compared KGML-SM trained with and without dataset filtering. As shown in \autoref{tab:filter_ablation}, filtering substantially improved model performance across multiple years. This demonstrates that eliminating unreliable simulation samples is critical for constructing a robust pretraining dataset and enhancing generalization to county-level observations.

\begin{table}[htbp]
  \centering
  \caption{Effect of filtering the simulated dataset on model performance (RMSE and R$^2$ for 2019–2023).}
  \scalebox{0.7}{
    \begin{tabular}{lcccccccccc}
    \toprule
    \multirow{2}{*}{\textbf{Method}} 
      & \multicolumn{2}{c}{2019} 
      & \multicolumn{2}{c}{2020} 
      & \multicolumn{2}{c}{2021} 
      & \multicolumn{2}{c}{2022} 
      & \multicolumn{2}{c}{2023} \\
    \cmidrule(lr){2-3}\cmidrule(lr){4-5}\cmidrule(lr){6-7}\cmidrule(lr){8-9}\cmidrule(lr){10-11}
      & RMSE & R$^2$ 
      & RMSE & R$^2$ 
      & RMSE & R$^2$ 
      & RMSE & R$^2$ 
      & RMSE & R$^2$ \\
    \midrule
    Unfiltered & 1.120 & 0.705 & 1.010 & 0.768 & 1.198 & 0.807 & 1.305 & 0.775 & 1.225 & 0.760 \\
    Filtered (KGML-SM)  & \textbf{0.964} & \textbf{0.741} & \textbf{0.980} & \textbf{0.792} & \textbf{1.104} & \textbf{0.836} & \textbf{1.085} & \textbf{0.837} & \textbf{1.071} & \textbf{0.807} \\
    \bottomrule
    \end{tabular}
  }
  \label{tab:filter_ablation}
\end{table}

\subsection{Prediction error spatialization and model bias}
\label{Prediction error spatialization and model bias}

\begin{figure*}[htbp]
\centering  %图片全局居中
\includegraphics[width=1\textwidth]{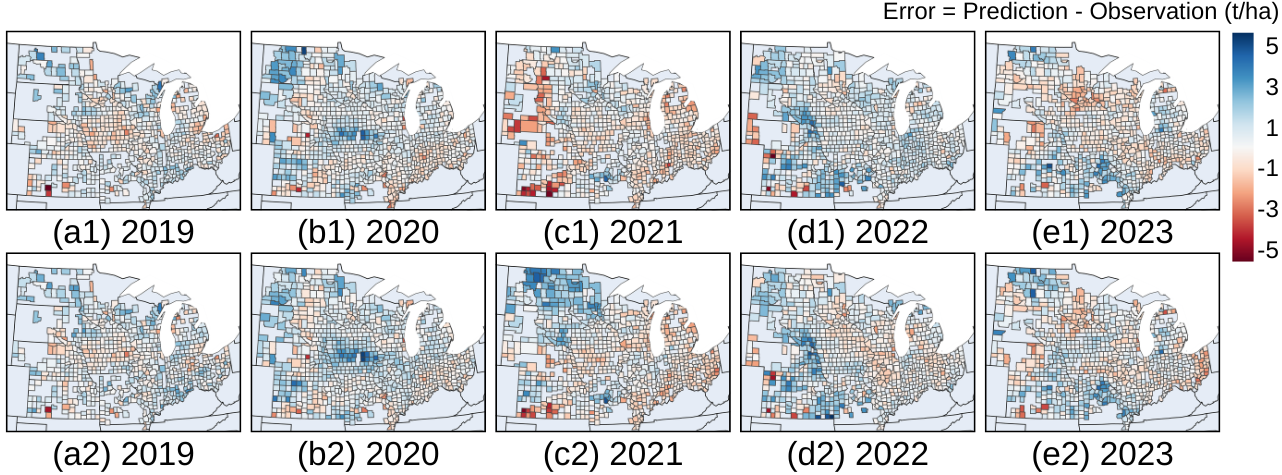}
\caption{The error map for (1) KGML-SM and (2) RF model from (a) 2019 to (e) 2023, with blue indicating overestimation.}
\label{fig:soilatt_att_err_map}
\end{figure*}

 In \autoref{fig:soilatt_att_err_map}, we present error maps of prediction results from 2019 to 2023 for the two best-performing models: KGML-SM and the RF baseline. Across all five years, KGML-SM consistently reduced overestimation compared with RF. In 2019, while overestimation in drought-affected areas of North Dakota was not substantially alleviated, several eastern states showed reductions. This is consistent with the fact that yield losses in 2019 were mainly due to delayed planting caused by prevented planting \citep{fb2019preventplant,usda2019preventplant}, rather than drought. In 2020, although the derecho storm \citep{derecho} in Iowa also posed challenges for KGML-SM, the RF model exhibited much more severe overestimation, indicating that KGML-SM is effective not only under drought conditions but also in mitigating overestimation associated with other natural hazards. The improvements were particularly pronounced in 2021 and 2022, when RF showed widespread overestimation across the northwestern Corn Belt, corresponding to regions of drought-induced yield reduction in 2021, and across Kansas and Nebraska in 2022. In 2023, KGML-SM also reduced overestimation in Minnesota, further demonstrating its robustness. These results show the robustness of KGML-SM in mitigating systematic yield overestimation across diverse climatic conditions.

% \clearpage

% \clearpage

\begin{figure*}[htbp]
\centering  %图片全局居中
\includegraphics[width=1\textwidth]{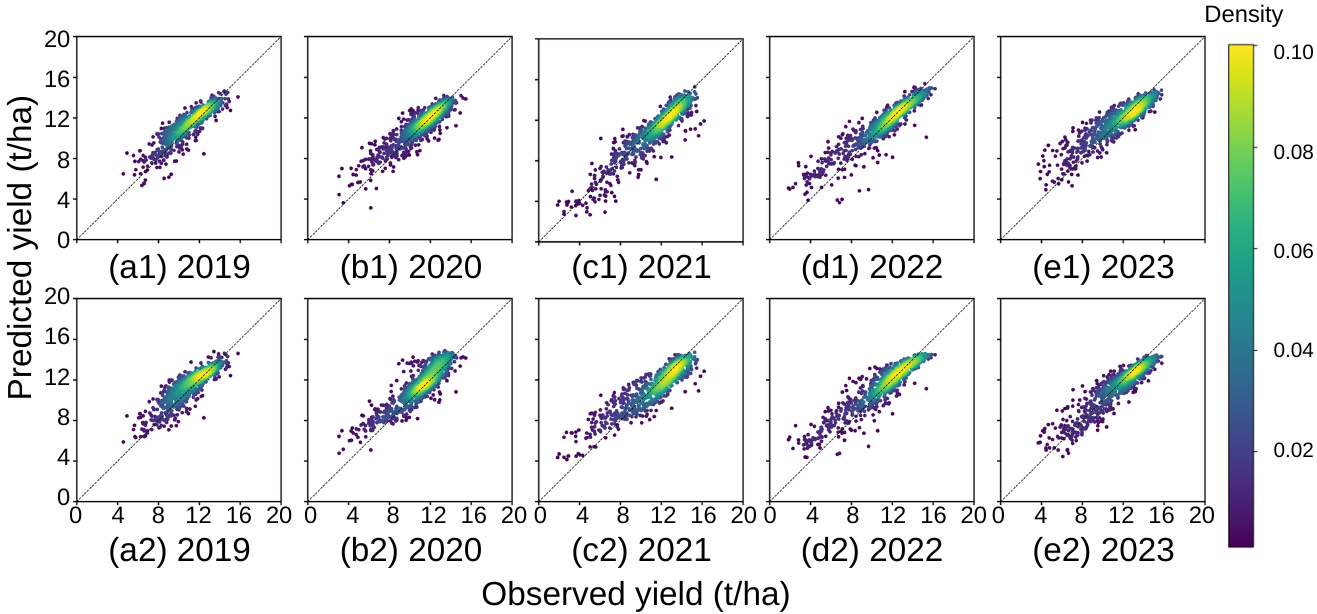}
\caption{Scatter plots of predicted versus observed county-level corn yields for KGML-SM (top row) and the RF model (bottom row) from 2019 to 2023. Panels (a1)–(e1) correspond to KGML-SM predictions for 2019–2023, while panels (a2)–(e2) correspond to RF predictions for the same years. The horizontal axis represents the reported (observed) yield (t/ha), and the vertical axis represents the model-predicted yield (t/ha). The color scale indicates the density of counties, and the dashed line shows the 1:1 reference line for comparison.}
\label{fig:soilatt_att_scatter}
\end{figure*}

To analyze model prediction performance, we generated scatter plots for the KGML-SM model and the RF model  (\autoref{fig:soilatt_att_scatter}). These plots helped visualize the relationship between observed and predicted values, revealing patterns of overestimation, underestimation, and potential prediction biases across different years. In 2019 (\autoref{fig:soilatt_att_scatter}(a)), the KGML-SM model presented a noticeably narrower distribution, indicating a lower spread in prediction errors. In 2020 (\autoref{fig:soilatt_att_scatter}(b)), the predictions of the KGML-SM model were noticeably more concentrated along the diagonal and exhibited symmetry on both sides, whereas the RF model produced more dispersed predictions in high-yield regions. In 2021 (\autoref{fig:soilatt_att_scatter}(c)), the RF model exhibited prediction collapse, where certain observed values corresponded to nearly identical predicted values, likely due to overfitting or insufficient variability in learned representations. In 2022 (\autoref{fig:soilatt_att_scatter}(d)) and 2023 (\autoref{fig:soilatt_att_scatter}(e)), the KGML-SM model maintained a narrower and more concentrated prediction distribution. This comparison highlighted the advantage of the KGML-SM model in mitigating prediction collapse and improving overall robustness across different years.

% \clearpage

\section{Discussion}
\label{Discussion}

In this section, we explore the role of soil moisture in model prediction from four questions:

\begin{itemize}
    \item (\autoref{Spatial influence of soil moisture on model prediction}) How did soil moisture influence model prediction spatially? 
    
    \item (\autoref{Temporal role of soil moisture during the corn growth season}) How did soil moisture affect model performance throughout the corn growth season? 
    
    \item (\autoref{Statistical impact of soil moisture in drought and non-drought regions}) how did soil moisture contribute to model prediction in drought and non-drought regions? 

    \item (\autoref{Interpreting corn yield prediction errors via soil moisture prediction anomalies}) How to interpret the observed inaccuracies in corn yield predictions based on soil moisture?
\end{itemize}

% sm attention in map, show spatial analysis

\subsection{Spatial influence of soil moisture on model prediction}
\label{Spatial influence of soil moisture on model prediction}

\begin{figure*}[htbp]
\centering  %图片全局居中
\includegraphics[width=1\textwidth]{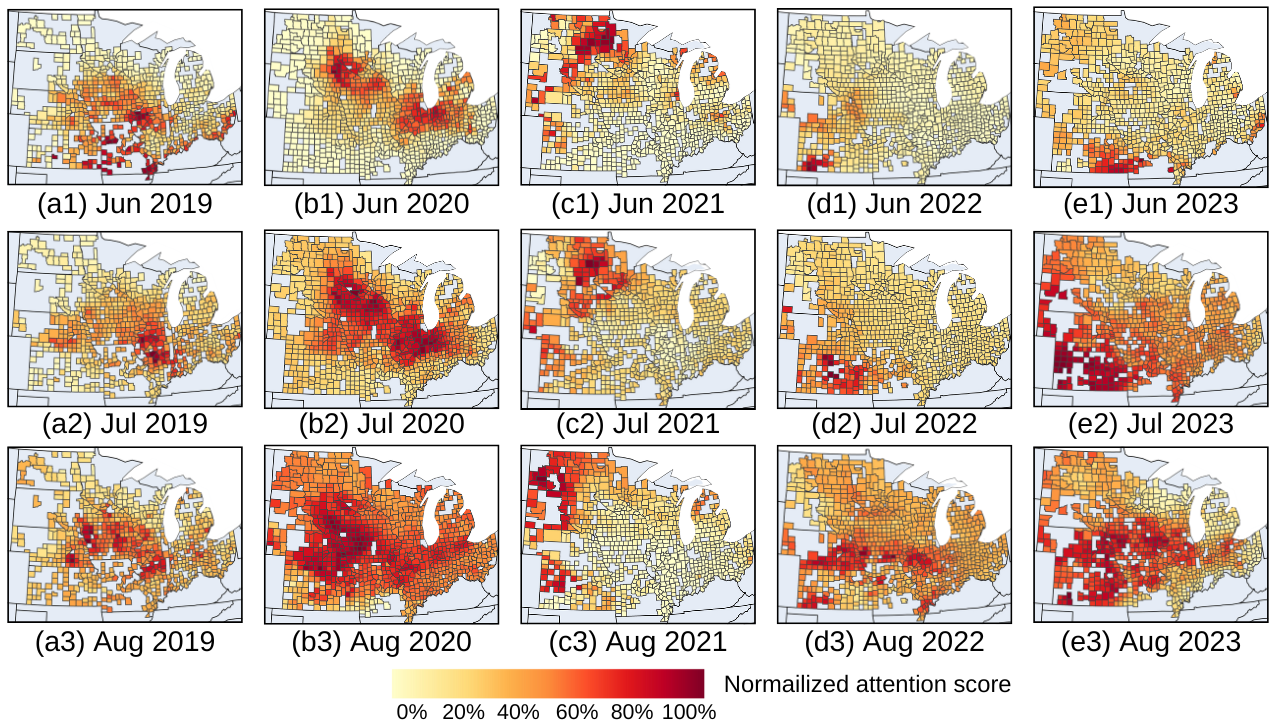}
\caption{The attention visualization of soil moisture on corn yield prediction in different regions. The corn growth period was segmented into three stages, covering (1) June to (3) August, for the years (a) 2019 to (e) 2023. The attention visualization was computed by first calculating the attention values for all counties. These values were then normalized based on the maximum value in this year to ensure a consistent color scale, making the distribution more visually interpretable.}
\label{fig:sm_att_map}
\end{figure*}

To answer the first question, we visualized the attention scores of soil moisture across twelve states in the U.S. Corn Belt from June to August over the years 2019 to 2023 (\autoref{fig:sm_att_map}). The attention scores indicated the relative importance assigned to soil moisture by the model in different regions, with higher scores suggesting a stronger influence on yield prediction. The attention map highlighted how the model's reliance on soil moisture varied across different growth stages and drought conditions, allowing us to assess whether soil moisture has a greater impact on the model in drought-affected areas. To enhance visualization, we normalized attention values within each year. Consequently, the analysis focused on attention trends across regions within the same year, while cross-year comparisons were not meaningful due to the normalization.

\clearpage

% \clearpage

In June, during the early growth period, the attention scores for soil moisture were generally lower across all five years, This might have been because the temperature was not too high and drought conditions were not severe at this time, leading to a less pronounced correlation between soil moisture and corn yield. In July, attention scores increased, particularly in drought-affected areas, as soil moisture became more influential during active growth and vegetative stages. By August, attention scores peaked in drought-affected states, aligning with the critical reproductive phase of corn when adequate soil moisture was essential for kernel development.

From 2019 to 2023, the attention distribution of soil moisture in corn yield prediction exhibited noticeable variations. In early 2019 (\autoref{fig:sm_att_map}(a1)), attention concentrated in Iowa, which does not align with the drought and yield reduction observed in North Dakota. An explanation is the widespread prevented planting reported that year, with the highest levels occurring in parts of eastern South Dakota, northwest Ohio, northeast Illinois, southwest Minnesota, and along the Mississippi \citep{fb2019preventplant, usda2019preventplant}, which disrupted normal planting schedules and shifted yield risks beyond drought-affected areas. During 2020 (\autoref{fig:sm_att_map}(b)), high-attention regions were initially concentrated in Iowa and Nebraska, later expanding to adjacent areas. By 2021 (\autoref{fig:sm_att_map}(c)), attention intensified over North Dakota, South Dakota, Minnesota, and Nebraska. In 2022 (\autoref{fig:sm_att_map}(d)), high-attention areas were primarily located in Nebraska, Kansas, South Dakota, and Iowa. In 2023 (\autoref{fig:sm_att_map}(e1)), attention initially focused on Kansas, expanding thereafter across a majority of states. Ultimately, significant attention emerged in Nebraska, Kansas, Minnesota, Iowa, Wisconsin, and Missouri, indicative of substantial drought conditions (\autoref{fig:sm_att_map}(e3)). Overall, the period from 2019 to 2023 exhibited a trend of drought conditions and corresponding attention expanding towards the central United States, consistent with actual observations. The reason for the incomplete consistency between the two could be that although drought was present in the region, it did not severely impact corn yield. Alternatively, even without drought in a particular area, other factors, such as the storm or flooding, might have affected corn yield. This analysis underscored the dynamic role of soil moisture in model prediction, with its importance intensifying during key growth stages and under severe drought conditions.

% average attention of sm, VIs, weather over time. line chart
  
\subsection{Temporal role of soil moisture during the corn growth season}
\label{Temporal role of soil moisture during the corn growth season}

\begin{figure*}[htbp]
\centering  %图片全局居中
\includegraphics[width=0.5\textwidth]{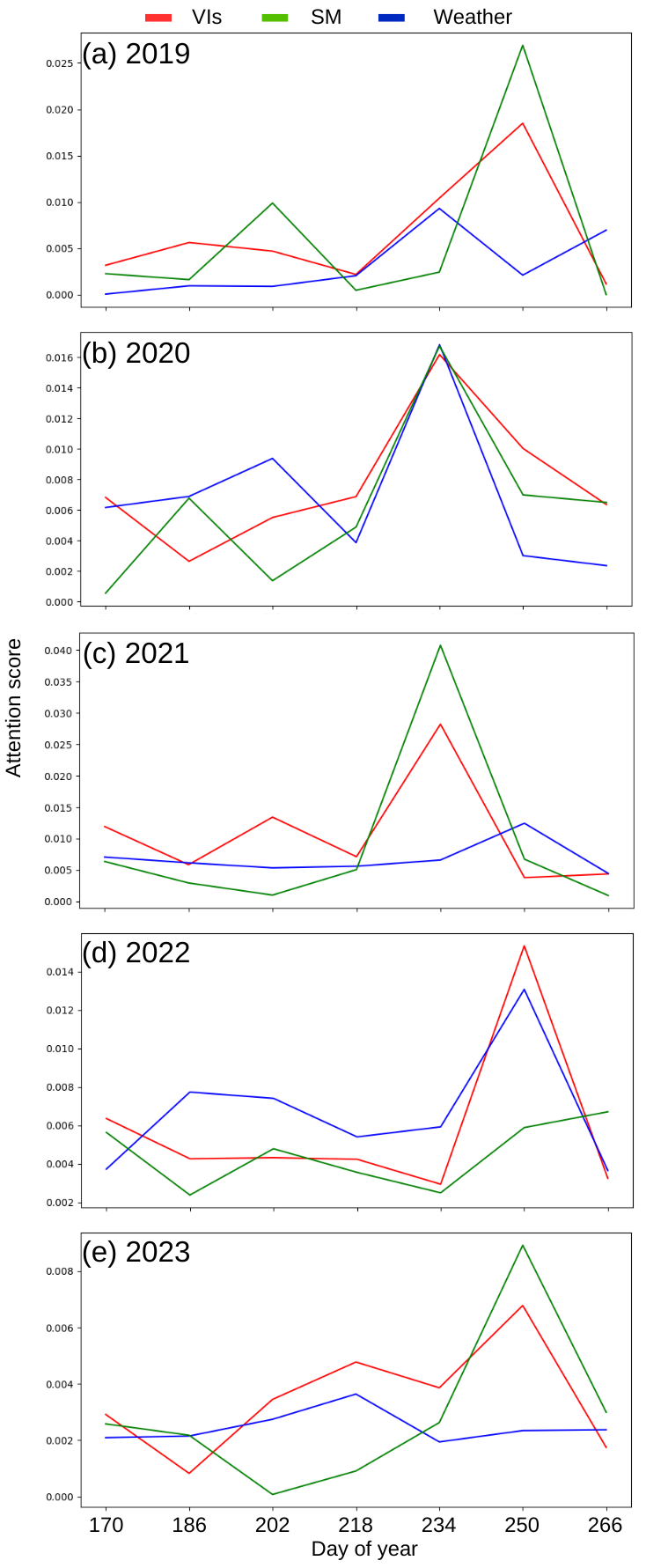}
\caption{5 year attention values of different feature types in the time series. The attention visualization is computed by first calculating the attention values for all features across all time points. Then, for each time point, we average the attention values across all features within the same category (VIs, Weather, and SM).}
\label{fig:line_chart}
\end{figure*}

To answer the second question and illustrate the impact of soil moisture at different stages of the corn growing season, we visualized the attention of three feature types—VIs, weather, and soil moisture—at 16-day intervals from June to September in drought-affected states (\autoref{fig:line_chart}). We found that the VIs had the highest influence on the model around August, which aligned with findings from previous study \citep{feature_importance}. In this period, VIs showed the strongest correlation with corn yield as this period corresponded to the vegetative and reproductive growth stages, during which crop health and biomass accumulation significantly impact final yield \citep{ndvi_correlation_1}. High VIs in this timeframe indicate optimal chlorophyll content, canopy development, and water availability, which are critical for photosynthesis and grain formation \citep{ndvi_correlation_2}. Additionally, we found that weather data showed a significant increase in attention around July in 2020 (\autoref{fig:line_chart}(b)) and 2022 (\autoref{fig:line_chart}(d)), and the periods with noticeable attention spikes aligned with the trends of VIs. This phenomenon was more pronounced in soil moisture, as soil moisture attention showed a strong correlation with VIs in all years except 2022. In 2022, a slight increase in soil moisture attention could still be observed near the VIs peak. This might be because VIs during these periods are closely related to certain weather data and soil moisture data. For example, NDWI increases with higher PPT because more rainfall enhances soil moisture and plant water content \citep{ndwi_ppt}. This indicated that our attention mechanism effectively captured feature importance over time dimension.

\clearpage

% box plot show attention of all the county in time dimension, drought and non-drought area two box

\subsection{Statistical impact of soil moisture in drought and non-drought regions}
\label{Statistical impact of soil moisture in drought and non-drought regions}

\begin{figure*}[htbp]
\centering  %图片全局居中
\includegraphics[width=1\textwidth]{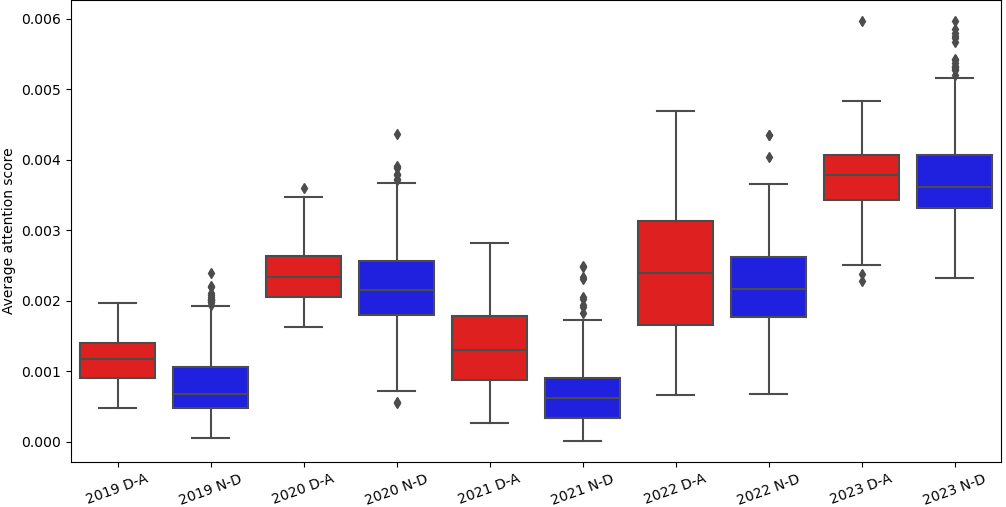}
\caption{The boxplot of soil moisture attention in drought-affected areas (D$-$A) and non-drought areas (N$-$D) across all counties.}
\label{fig:boxplot_arid_nonarid}
\end{figure*}

To answer the third question, the box plot was used to illustrate the distribution of soil moisture attention across all counties from 2019 to 2023 (\autoref{fig:boxplot_arid_nonarid}). Each point in this box plot represents a county, displaying the comparison of attention between red-marked drought-affected areas and blue-marked non-drought areas from 2019 to 2023 based on \autoref{tab:drought_reduced_yield_stat}. Across all five years, drought-affected areas exhibited fewer outliers compared to non-drought areas, suggesting that the model's attention to soil moisture is more stable in drought-affected regions. Additionally, in all years, the median of soil moisture attention in drought-affected areas was consistently higher than in non-drought areas, indicating that soil moisture had a greater impact on the model in these areas.

% \clearpage

\subsection{Interpreting corn yield prediction errors via soil moisture prediction anomalies}
\label{Interpreting corn yield prediction errors via soil moisture prediction anomalies}

\begin{figure*}[htbp]
\centering  %图片全局居中
\includegraphics[width=1\textwidth]{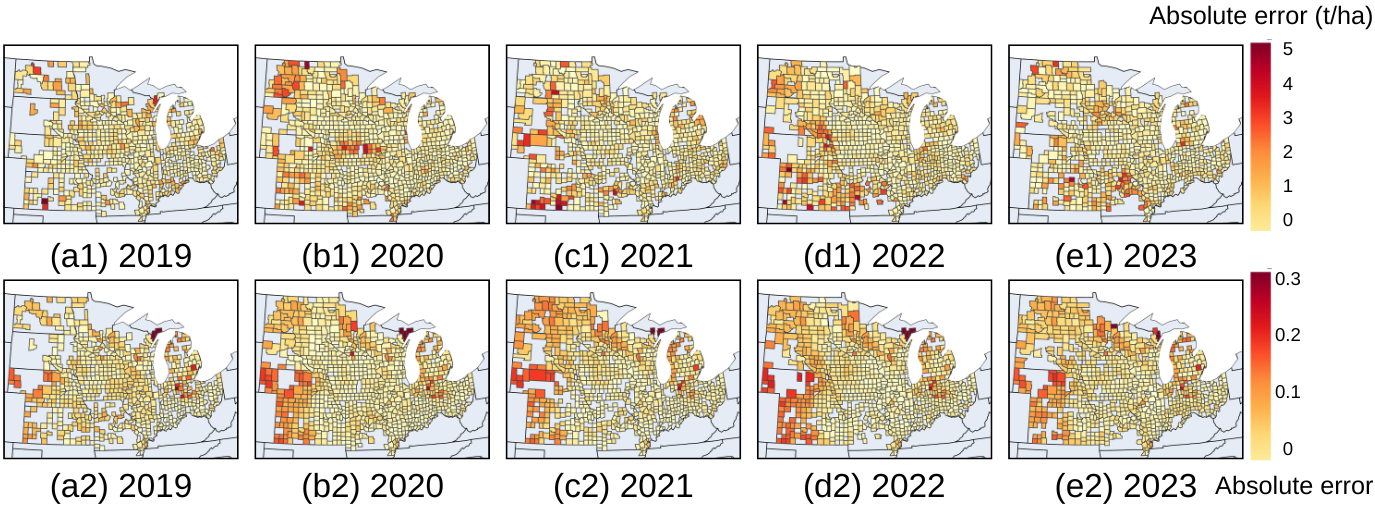}
\caption{The absolute error map for (1) corn yield prediction and (2) soil moisture prediction from (a) 2019 to (e) 2023.}
\label{fig:sm_yield_abs_err_map}
\end{figure*}

To answer the fourth question, we plotted the absolute error map of both soil moisture and corn yield predictions (\autoref{fig:sm_yield_abs_err_map}). In our modeling process, we treated soil moisture as an intermediate variable to guide the corn yield output, so when corn yield prediction was inaccurate, the corresponding soil moisture  might also have exhibited anomalies. This analysis aims to interpret inaccurate corn yield predictions.

In 2019 (\autoref{fig:sm_yield_abs_err_map}(a)), the absolute error map shows that corn yield predictions in North Dakota exhibited relatively large errors, while the corresponding soil moisture predictions remained accurate. This inconsistency suggests that yield errors were not primarily driven by soil moisture misrepresentation but were instead likely associated with large-scale prevented planting events in 2019 \citep{fb2019preventplant, usda2019preventplant}, which cannot be fully explained by soil moisture dynamics. In 2020 (\autoref{fig:sm_yield_abs_err_map}(b)), inaccurate corn yield prediction in Iowa was primarily attributable to the derecho storm \citep{derecho}, which soil moisture could not reflect. In 2021 and 2022 (\autoref{fig:sm_yield_abs_err_map}(c–d)), the model exhibited large corn yield prediction errors in drought-affected regions such as North and South Dakota, Nebraska, and Kansas. These yield errors coincided with inaccurate soil moisture predictions in the same areas, suggesting that misrepresentation of soil moisture dynamics under drought conditions was a key contributor to yield prediction inaccuracies. In 2023 (\autoref{fig:sm_yield_abs_err_map}(e)), prediction errors were observed in Missouri and Minnesota. In Missouri, however, the corresponding soil moisture predictions remained relatively accurate. Field reports documented damaging winds and hail in late June and early July, particularly around Mooresville, Rockport, and Garden City, which caused root lodging and crop damage \citep{mo_anomaly}. Because such mechanical damage is not captured by standard soil moisture data, it likely contributed to the yield prediction errors in this region.

Overall, these results indicate that a significant portion of corn yield prediction error is linked to soil moisture prediction error, underscoring the central role of soil moisture in crop growth. By explicitly modeling soil moisture as an intermediate variable, KGML-SM provides a diagnostic layer of interpretability: yield prediction errors can be traced to misrepresentation of soil moisture dynamics. This design not only improves robustness under drought conditions but also offers a transparent explanation for when and why yield predictions fail.

% \clearpage

% \clearpage

\subsection{Strengths, limitations, and future work}
\label{Strengths, limitations, and future work}

Our main contributions are as follows:

\begin{itemize}
    \item  This paper introduced the KGML-SM framework, which integrates process-based and ML models for corn yield prediction while explicitly incorporating the influence of soil moisture.
    \item  A drought-aware loss function was designed to enhance model performance under drought conditions and mitigate overestimation.
    \item  Based on the relationship between drought, soil moisture, and corn yield prediction, we provided interpretability for the prediction errors of the KGML-SM model and offered directions for future model optimization.
\end{itemize}

While these contributions highlight the novelty and effectiveness of our approach, it is equally important to recognize its limitations and potential directions for future research. First, the study relies on SMAP soil moisture data and APSIM simulations, which are limited in temporal coverage and in the diversity of spatial resolution. Incorporating multi-scale data, such as finer field-level observations and broader regional products, could improve the model’s generalization across heterogeneous environments. Second, the drought-aware loss function is designed to mitigate overestimation under drought conditions, but it does not address other extreme events such as floods or heatwaves. Extending the loss function to include these scenarios would make the framework more comprehensive. Finally, the experiments are limited to the U.S. Corn Belt, and testing the KGML-SM framework in other regions and on different crops will be essential to evaluate its broader applicability. Addressing these limitations will further improve the robustness and generalizability of KGML-SM, ultimately contributing to more reliable and interpretable crop yield prediction under diverse environmental conditions.

\section{Conclusion}
\label{Conclusion}

In this study, we propose the KGML-SM model, where the W2S encoder is designed to capture the influence of weather on soil moisture, and the attention module is employed to weight different input features for final corn yield prediction. To address the issue that drought conditions often lead to yield overestimation, we introduce a drought-aware loss function to mitigate this problem in drought-affected regions. We construct both an APSIM field-level dataset and a GEE county-level dataset, learning the corn growth process by pretraining KGML-SM on the APSIM field-level dataset and then finetuning it on the GEE county-level dataset. Our analysis covers 12 states in the U.S. Corn Belt to investigate the impact of soil moisture on corn yield prediction. The proposed method consistently outperforms baseline models across multiple test years. Furthermore, we study the spatial and temporal influence of soil moisture through attention visualization, revealing when and where the model places greater focus on soil moisture. Finally, based on the relationship between soil moisture and corn yield prediction, we investigate the causes of prediction inaccuracies and provide explanations. In future work, we aim to apply transfer learning techniques to adapt models trained on well-studied regions with abundant simulated data to regions with limited data availability.

\section*{Funding}

This work was supported by the United States Department of Agriculture (USDA) National Institute of Food and Agriculture (NIFA) Agriculture and Food Research Initiative Foundational Program (Award No. 2022-67021-36468); and the USDA NIFA Hatch Project (Accession No. 7005141).

\clearpage

\bibliographystyle{elsarticle-harv} 
\bibliography{cas-refs}

\end{document}